\documentclass[11pt]{article}

\usepackage[final]{acl}

\usepackage{times}
\usepackage{latexsym}
\usepackage{booktabs}
\usepackage{multirow}
\usepackage{graphicx}
\usepackage{amsmath}
\usepackage{amssymb}
\usepackage{mathtools}
\usepackage{hyperref}
\usepackage{multicol}
\usepackage{multirow}
\usepackage{subcaption}
\usepackage[noabbrev,capitalize,nameinlink]{cleveref}
\usepackage{xcolor}

\newcommand{\up}[1]{\textcolor{blue}{$\uparrow${#1}}}
\newcommand{\down}[1]{\textcolor{red}{$\downarrow${#1}}}
\newcommand{\eurollm}{{\tt EuroLLM-9B}}
\newcommand{\aya}{{\tt aya-3B}}
\DeclareMathOperator*{\argmax}{arg\,max}
\DeclareMathOperator*{\argmin}{arg\,min}
\usepackage[T1]{fontenc}
\usepackage[utf8]{inputenc}
\usepackage{microtype}
\usepackage{inconsolata}
\usepackage{graphicx}
\usepackage{todonotes}
\usepackage{tikz}
\usepackage{tikz-dependency}
\newcommand*\circled[1]{\tikz[baseline=(char.base)]{
            \node[shape=circle,draw,inner sep=.6pt] (char) {#1};}}

\newcommand{\croco}{\textsc{CroCo}}

\title{\textsc{CroCo}: Cross-Lingual Contrastive Preference Tuning on Self-Generations}

\author{Mike Zhang\textsuperscript{{$\diamond$}{$\ddagger$}{$\dagger$}} \hspace{2em} Ali Basirat\textsuperscript{$\ddagger$} \hspace{2em} Desmond Elliott\textsuperscript{{$\diamond$}{$\dagger$}}\\
  \textsuperscript{$\diamond$}Department of Computer Science (DIKU), University of Copenhagen\\
  \textsuperscript{$\ddagger$}Centre for Language Technology (CST), University of Copenhagen \\
  \textsuperscript{$\dagger$}Pioneer Centre for Artificial Intelligence\\
  \small{
   \textbf{Correspondence:} \href{mailto:mike.zhang@di.ku.dk}{mike.zhang@di.ku.dk}
 }\\}

\begin{document}
\maketitle
\begin{abstract}
Prior work establishes that controlled contrastiveness between self-generated responses from large language models, set via reward scores, improves downstream preference tuning in English. We extend this method to multiple languages and evaluate two models
across a total of 14 high and low-resource languages on a diverse set of tasks. 
Our central finding is that \textbf{cro}ss-lingual \textbf{co}ntrastive preference tuning on self-generations (\croco{}) transfers without language-specific preference annotation. 
A reward model trained on English preferences (atop a multilingual base) produces useful within-language rankings across most languages, and pairing in either a monolingual or multilingual setting improves over each model on the majority of setups while preventing the catastrophic forgetting of supervised fine-tuning. 
We observe that the gains require on-policy data. Off-policy responses reduce the benefit and online preference optimization fails to improve over the offline variant. 
Specifically, on structured tasks, our method matches or exceeds the base in 6/7 languages for \eurollm{} and 4/7 settings for \aya{}. 
On open-ended generation, evaluated by two judges, both tuned models win 28/30 times against their respective base across 15 evaluated languages (high and low-resource). Overall, we show promising directions for multilingual preference tuning using self-generations.\footnote{The code is publicly available at \url{https://github.com/jjzha/CroCo}.}
\end{abstract}

\section{Introduction}

Aligning large language models (LLMs) with human preferences is the standard final stage of post-training, and Direct Preference Optimization~\citep[DPO;][]{rafailov2023direct} is one of the dominant approaches. Recently, DPO has been applied to self-generated samples rather than human preferences~\cite{guo2024directlanguagemodelalignment,xiao-etal-2025-finding}: a policy model is paired with a reward model (RM) that scores its on-policy responses to build preference pairs of \emph{chosen} and \emph{rejected} completions. Similarly, recent work has shifted attention from the optimizer to the data: \citet{pan2025what} show that chosen-response quality dominates downstream performance, \citet{gengdelta2025} establish that the \emph{relative} quality gap drives improvement, and \citet{xiao-etal-2025-finding} identify a ``sweet spot'' in which the rejected response is sampled near a specific quartile of the reward distribution rather than at the minimum. These findings are exclusively in English.

\begin{figure}[t]
    \centering
    \includegraphics[width=\linewidth]{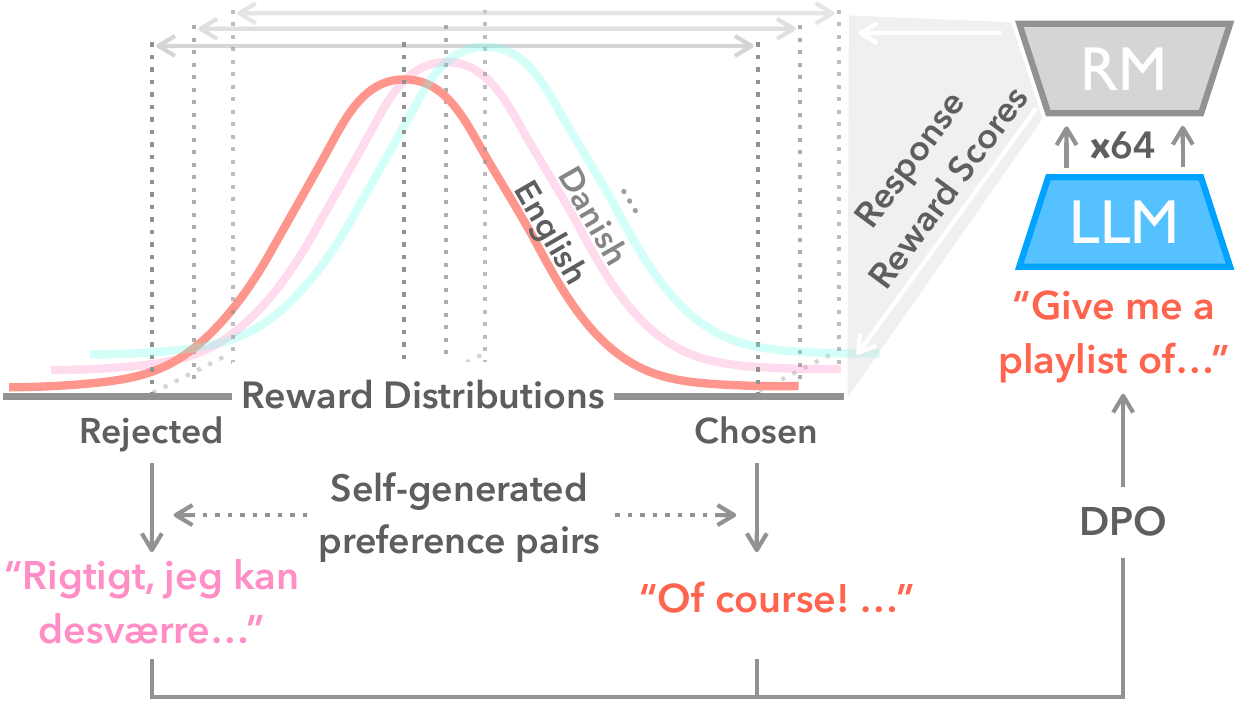}
    \caption{\textbf{Setup.} An LLM generates 64 responses per prompt per language; an external off-the-shelf RM scores these and we sample specific quartiles to construct \emph{contrastive} preference pairs.}
    \label{fig:setup}
\end{figure}

Extending preference tuning beyond English raises open questions. Prior multilingual work relies on translation-based preference signals~\citep{she-etal-2024-mapo}, exploits the English/non-English capability gap as an implicit reward~\citep{yang2025language, yang-etal-2025-implicit}, or reweights the DPO loss for noisy multilingual pairs~\citep{pokharel-etal-2025-capo}. None of these establishes whether reward-distribution-based pair construction itself transfers across languages. We therefore ask: \emph{Does contrastive preference tuning on self-generations transfer to a multilingual setting without language-specific preference annotation?} We examine this across monolingual and multilingual training regimes and two post-trained models at different scales (3B and 9B parameters).

\paragraph{Hypothesis.} We posit that contrastive preference tuning transfers cross-lingually, because the DPO objective depends on the relative reward gap rather than absolute calibration. Consistent \emph{within}-language ranking suffices despite cross-lingual miscalibration. This predicts that (i) an English-only RM --- built atop a multilingual base, as is standard for open RMs (e.g., \citealp{liu2026skyworkrewardv2scalingpreferencedata}) --- suffices for multilingual tuning when scored on within-language samples, removing the need for per-language annotation, and (ii) on-policy data matters more than generator quality, since the contrastive signal is informative only when paired responses come from the policy's own distribution.

\paragraph{Contributions.} \circled{1} Contrastive preference tuning transfers cross-lingually and across models: DPO on self-generations outperforms SFT baselines and existing multilingual preference-tuning methods~\citep{she-etal-2024-mapo, yang-etal-2025-implicit}, while standard SFT causes catastrophic forgetting in both models. \circled{2} Multilingual preference tuning does not require multilingual preference annotation: an English-only RM (atop a multilingual base) drives consistent gains across most languages, and joint multilingual training matches or exceeds monolingual training for both models. \circled{3} The method improves both structured and open-ended evaluation: multilingual {\tt Paired} DPO matches or exceeds the base in 6/7 languages for \eurollm{} and 4/7 settings for \aya{} on EuroEval, and both DPO-tuned models beat their base in all 11 evaluated languages on m-ArenaHard 2.1 using \texttt{Qwen3.6-35B-A3B} as the judge. \circled{4} Ablations on translation, prompt language, and on-policy vs.\ off-policy data confirm hypothesis (ii) and isolate which design choices are crucial, in line with \citet{tajwar2024preference} and \citet{shenfeld2026self}.

\section{Problem Formulation}
\label{sec:background}
\paragraph{Preference Tuning.}
Let $\pi_\theta$ be a policy language model parameterized by $\theta$, and $\pi_{\mathrm{ref}}$ a frozen reference model. Given a prompt $x$ and a preference pair $(y_c, y_r)$, where $y_c$ is \emph{chosen} over the \emph{rejected} $y_r$, DPO~\citep{rafailov2023direct} minimizes
\begin{equation}
    \mathcal{L}_{\mathrm{DPO}}(\theta) = - \mathop{\mathbb{E}}_{(x, y_c, y_r) \sim \mathcal{D}} \Big[ \log \sigma \big( \Delta r_\theta \big) \Big],
\end{equation}

where $\Delta r_\theta \coloneqq r_\theta(x, y_c) - r_\theta(x, y_r)$ is the reward margin, $r_\theta(x, y) \coloneqq \beta \log \bigl( \pi_\theta(y \mid x) / \pi_{\mathrm{ref}}(y \mid x) \bigr)$ is the implicit reward, and $\sigma(\cdot)$ is the sigmoid. The quality of the dataset $\mathcal{D} = \{(x^{(i)}, y_c^{(i)}, y_r^{(i)})\}_{i=1}^{N}$ is central to downstream performance.

\paragraph{Contrastive Preference Pairs.}
Following~\citet{xiao-etal-2025-finding}, we build $\mathcal{D}$ via on-policy self-generation. For each prompt $x$, the policy generates $K$ candidates $\mathcal{Y}_x = \{y^{(k)}\}_{k=1}^{K}$, each scored by an external reward model $R \colon \mathcal{X} \times \mathcal{Y} \to \mathbb{R}$. With $\mu_x, \sigma_x$ the mean and standard deviation of $\{R(x, y^{(k)})\}_{k=1}^{K}$, a preference pair is formed as

\begin{equation}
\begin{aligned}
    y_c &\;=\; \argmax_{y \in \mathcal{Y}_x}\; R(x, y), \\
    y_r &\;=\; \argmin_{y \in \mathcal{Y}_x}\; \bigl| R(x, y) - (\mu_x - 2\sigma_x) \bigr|.
\end{aligned}
\label{eq:sweet_spot}
\end{equation}

In other words, rather than targeting the lowest-scoring candidate, $y_r$ is selected as the sample in $\mathcal{Y}_x$ whose reward is nearest to $\mu_x - 2\sigma_x$, inducing a controlled level of contrastiveness between $y_c$ and $y_r$. We show samples from each region of the reward distribution in~\cref{app:reward_samples}.

\paragraph{Multilingual Extension.}
Prior work establishes this construction only for English; we extend it to target languages $\mathcal{L} = \{\ell_1, \dots, \ell_L\}$. Given an English prompt set $\mathcal{X}_{\mathrm{eng}}$, we obtain parallel prompts $\mathcal{X}_\ell$ for each $\ell$ via machine translation. For every $(x, \ell)$, the policy generates $K$ responses conditioned on the $\ell$-language prompt, yielding a language-specific dataset $\mathcal{D}_\ell$. We study two settings: (1)~\textbf{Monolingual}, tuning on each $\mathcal{D}_\ell$ independently, and (2)~\textbf{Multilingual}, tuning jointly on $\mathcal{D} = \bigcup_{\ell \in \mathcal{L}} \mathcal{D}_\ell$. We use two models of different scales (3B/9B) to test robustness to model size.

\section{Experimental Setup}

\subsection{Data}\label{sec:data}

\begin{figure}[t]
    \centering
    \includegraphics[width=\linewidth]{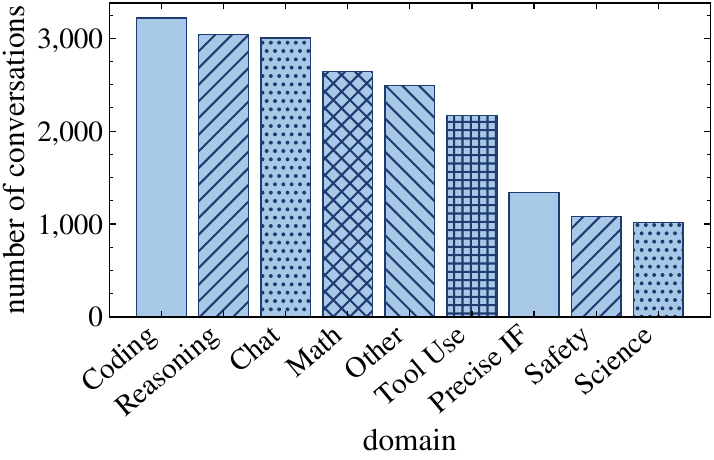}
    \caption{\textbf{Domain distribution of Dolci-Instruct-SFT.} Our 20K stratified sample covers nine task domains, with coding, reasoning, chat, and math accounting for the bulk of instances.}
    \label{fig:domain}
\end{figure}

\begin{figure}[t]
    \centering
    \includegraphics[width=.8\linewidth]{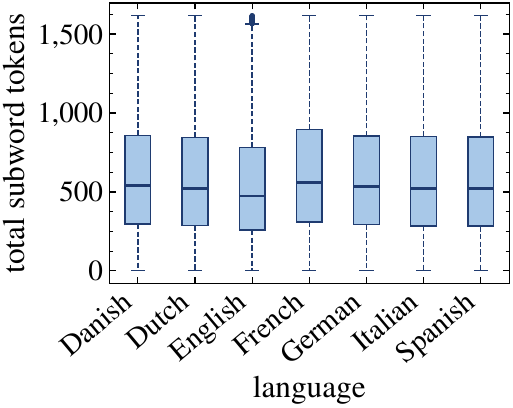}
    \caption{\textbf{Subword-token length distribution across languages.} We cap the 90th percentile at 1,616 tokens. Romance languages (French, Italian, Spanish) produce systematically longer translations than Germanic ones.}
    \label{fig:boxplot}
\end{figure}

We stratify 20K instances from Dolci-Instruct-SFT, the instruction tuning corpus used to train OLMo3~\citep{olmo2025olmo3}; the sampled domain distribution is shown in~\cref{fig:domain}. We translate the English data into six European languages: Danish (\texttt{dan}), Dutch (\texttt{nld}), French (\texttt{fra}), German (\texttt{deu}), Italian (\texttt{ita}), and Spanish (\texttt{spa}), using \texttt{TranslateGemma-27B}~\citep{finkelstein2026translategemmatechnicalreport}. Token-length statistics per language are reported in~\cref{fig:boxplot}.

Using \eurollm{}\footnote{\url{https://huggingface.co/utter-project/EuroLLM-9B-Instruct-2512}.}~\citep{ramos2026eurollm} or \aya{}\footnote{{\url{https://huggingface.co/CohereLabs/tiny-aya-global}}.}~\citep{salamanca2026tiny} as the on-policy model, we generate 64 responses per instance ($>60$ samples plateaus performance per~\citealp{xiao-etal-2025-finding}) at temperature $T=0.7$ for \eurollm{} and $T=0.1$ for \aya{}, producing 1.28M samples per language. Each is scored with \texttt{Skywork-Reward-V2-Qwen3-8B}~\citep{liu2024skyworkrewardbagtricksreward, liu2026skyworkrewardv2scalingpreferencedata}, an RM whose preference training is English-only but whose model (Qwen3-8B) is multilingual~\citep{yang2025qwen3}. We select this RM because English-preference-trained RMs of this kind transfer robustly across languages~\citep{wu-etal-2024-reuse, hong-etal-2025-cross} and because it ranks sixth on RewardBench 2.0~\citep{malik2025rewardbench}.\footnote{\url{https://huggingface.co/spaces/allenai/reward-bench}} Crucially, our hypothesis requires the RM to \emph{score} responses consistently within and across each target language. We show this happens qualitatively in~\cref{app:reward_dists}.

\paragraph{Training Data Construction.} We compare four construction strategies, in both monolingual and multilingual regimes, applied to both models:
    \circled{1} \textbf{In-Lang} / \circled{2} \textbf{All Lang (SFT)}: the translated in-language set, or the union across all languages, fine-tuned with standard SFT, without any preference signal.
    \circled{3} \textbf{Max-R (SFT)}: for each prompt, only the highest-scoring response is kept and SFT applied: a best-of-$K$ baseline that uses the reward signal but discards contrastiveness.
    \circled{4} \textbf{Paired (DPO)}: following \citet{xiao-etal-2025-finding}, we form preference pairs following 
    \cref{eq:sweet_spot}, and apply DPO.

We verify in~\cref{app:lang_selection} that the multilingual {\tt Paired} construction does not degenerate into selecting English as chosen and a non-English language as rejected, but selects across all languages.

\subsection{Training}

We fine-tune with LoRA~\citep{hu2022lora} for all setups in TRL~\citep{vonwerra2020trl}.\footnote{We are aware of the gradient accumulation and CPU offloading bug found by~\citet{limozin2026sft} in SFT training using TRL; we detail in~\cref{app:params} how we are not affected.} For SFT, we train for 1 epoch with sequence length 4{,}096, global batch size 64, and learning rate $2 \times 10^{-4}$ (cosine schedule, 5\% warmup, weight decay $1 \times 10^{-2}$), optimizing the standard autoregressive cross-entropy loss over completions only.

For preference tuning, the policy $\pi_\theta$ also serves as the frozen reference $\pi_\mathrm{ref}$. We train for 1 epoch with learning rate $5 \times 10^{-6}$ (cosine schedule, 5\% warmup, weight decay $1 \times 10^{-2}$), $\beta = 0.1$, and the same batch size and sequence length as SFT. Full training details are in~\cref{app:params}.

\begin{table*}[t]
    \centering
    \small
    \setlength{\tabcolsep}{4pt}
        \begin{tabular}{ll ccc ccc ccc}
    \toprule
    & & \multicolumn{3}{c}{\tt Baselines} & \multicolumn{3}{c}{\tt Monolingual Post-training} & \multicolumn{3}{c}{\tt Multilingual Post-training} \\
    \cmidrule(lr){3-5} \cmidrule(lr){6-8} \cmidrule(lr){9-11}
    & & {\tt Base} & {\tt ICR} & {\tt MAPO} & {\tt In-lang} & {\tt Max-R} & {\tt Paired} & {\tt All Lang} & {\tt Max-R} & {\tt Paired} \\
    Language & model & & {\tt 8B} & {\tt 13B} & {\tt (SFT)} & {\tt (SFT)} & {\tt (DPO)} & {\tt (SFT)} & {\tt (SFT)} & {\tt (DPO)} \\
    \midrule
    \multirow{2}{*}{dan (7)} 
        & {\tt aya-3B}     & 48.4 & -          & -          & \down{4.0}     & \down{4.2}             & \up{0.1}       & \down{4.6}     & \down{7.4}     & \up{0.1} \\
        & {\tt EuroLLM-9B} & 65.0 & -          & -          & \down{5.3}     & \up{0.1}       & \up{1.0}       & \down{8.5}     & \up{1.3}       & \up{1.1} \\
    \cmidrule(l){2-11}
    \multirow{2}{*}{deu (4)} 
        & {\tt aya-3B}     & 42.4 & \down{1.7} & \down{0.7} & \down{4.2}     & \down{5.7}             & \down{0.1}     & \down{6.9}     & \down{7.6}     & \down{0.1} \\
        & {\tt EuroLLM-9B} & 47.4 & \down{6.7} & \down{5.7} & \up{0.4}       & \down{2.0}     & \up{1.0}       & \down{3.8}     & \down{0.9}     & \up{1.2} \\
    \cmidrule(l){2-11}
    \multirow{2}{*}{eng (5)} 
        & {\tt aya-3B}     & 54.1 & \down{1.0} & 0.0        & \down{2.8}     & \down{6.3}             & \up{0.2}       & \down{7.9}     & \down{7.0}     & \up{0.4} \\
        & {\tt EuroLLM-9B} & 56.8 & \down{3.6} & \down{2.6} & \down{0.1}     & \down{0.5}     & \down{0.4}     & \down{3.9}     & \up{0.2}       & \up{0.3} \\
    \cmidrule(l){2-11}
    \multirow{2}{*}{spa (4)} 
        & {\tt aya-3B}     & 42.8 & \down{1.0} & \up{2.3}   & \down{1.4}     & \down{3.0} & \down{0.1}     & \down{3.8}     & \down{4.5}     & \down{0.5} \\
        & {\tt EuroLLM-9B} & 51.8 & \down{10.0} & \down{6.7} & \down{2.9}     & \down{1.7}     & \down{1.3}     & \down{4.8}     & \down{2.1}     & \up{0.9} \\
    \cmidrule(l){2-11}
    \multirow{2}{*}{fra (5)} 
        & {\tt aya-3B}     & 45.9 & \down{3.6} & \down{1.1} & \down{5.2}     & \down{6.8}             & \down{0.1}     & \down{7.2}     & \down{7.2}     & \down{0.3} \\
        & {\tt EuroLLM-9B} & 52.2 & \down{9.9} & \down{7.4} & \down{2.0}     & \down{1.3}     & \up{0.6}       & \down{6.6}     & \down{2.7}     & \up{0.8} \\
    \cmidrule(l){2-11}
    \multirow{2}{*}{ita (3)} 
        & {\tt aya-3B}     & 51.6 & -          & -          & \down{10.7}    & \down{10.6} & \up{0.4}       & \down{11.3}    & \down{10.5}    & \up{0.6} \\
        & {\tt EuroLLM-9B} & 54.3 & -          & -          & \down{7.6}     & \up{1.2}       & \up{3.6}       & \down{9.9}     & \up{5.0}       & \down{1.3} \\
    \cmidrule(l){2-11}
    \multirow{2}{*}{nld (4)} 
        & {\tt aya-3B}     & 58.5 & -          & -          & \down{6.0}     & \down{6.1}             & \down{0.1}     & \down{10.8}    & \down{7.3}     & \up{0.3} \\
        & {\tt EuroLLM-9B} & 68.0 & -          & -          & \down{3.9}     & \down{1.7}     & \up{0.2}       & \down{6.5}     & \down{0.9}     & 0.0 \\
    \bottomrule
    \end{tabular}

    \caption{\textbf{Average EuroEval evaluation summarized by language, model, and tasks.} The {\tt Base} column shows the absolute aggregated EuroEval score for each model over three seeds. All other columns show the absolute difference from the model on the same row. ICR~\citep{yang-etal-2025-implicit} and MAPO~\citep{she-etal-2024-mapo} are independent baseline models with the parameter counts shown in their column headers. Number of datasets per language in parentheses. English uses the original Dolci SFT data. We show the exact numbers per dataset in \cref{app:exact_nums}.}
    \label{tab:main_results}
\end{table*}

\begin{table}[t]
    \centering
    \small
    \setlength{\tabcolsep}{3pt}
    \begin{tabular}{lcccc}
    \toprule
    & \tt Baseline & \multicolumn{2}{c}{\tt SFT} & \multicolumn{1}{c}{\tt DPO} \\
    \cmidrule(lr){2-2} \cmidrule(lr){3-4} \cmidrule(lr){5-5}
    Lang. & {\tt EuroLLM} & {\tt All Lang} & {\tt Max-R} & {\tt Paired} \\
    \midrule
    nor (5) & 54.3 & \down{5.7} & \up{0.1} & \up{0.5} \\
    por (2) & 47.4 & \down{8.2} & \up{0.1} & \up{0.4} \\
    swe (4) & 52.8 & \down{5.2} & \down{0.3} & \up{0.2} \\
    \bottomrule
    \end{tabular}
    \caption{\textbf{Cross-lingual generalization to held-out languages in EuroEval (Norwegian, Portuguese, Swedish).} Values are dataset-averaged absolute differences from the \eurollm{} baseline; the count of held-out datasets per language is in parentheses. {\tt Paired} DPO generalizes positively in all three held-out languages, while multilingual SFT degrades performance.}
    \label{tab:eurollm_crosslingual_results_avg}
\end{table}

\subsection{Evaluation}

We evaluate with EuroEval~\citep{smart2023scandeval, smart2024encoder}, a multilingual framework supporting all European languages. The suite comprises 32 datasets across the seven target languages (\texttt{dan}, \texttt{nld}, \texttt{eng}, \texttt{fra}, \texttt{deu}, \texttt{ita}, \texttt{spa}), covering reading comprehension, knowledge, commonsense reasoning, linguistic acceptability, and word-in-context tasks; full details are in~\cref{app:datasets}. For cross-lingual generalization analyses we additionally evaluate on Norwegian (\texttt{nor}), Portuguese (\texttt{por}), and Swedish (\texttt{swe}). For open-ended generation we use m-ArenaHard 2.1 (\cref{sec:arenahard}), where we evaluate on \texttt{dan}, \texttt{nld}, \texttt{eng}, \texttt{fra}, \texttt{deu}, \texttt{ita}, \texttt{spa}, Galician (\texttt{glg}), Irish (\texttt{gle}), Maltese (\texttt{mlt}), and Welsh (\texttt{cym}).

Additionally to complement structured prediction (i.e., EuroEval), we evaluate the models on m-ArenaHard 2.1~\citep{salamanca2026tiny}. This is a benchmark to evaluate open-ended generation acros languages. It is a multilingual extension of ArenaHard~\citep{pmlr-v267-li25h} covering English, German, Spanish, French, Italian, and Dutch, with 498 prompts per language across coding, creative writing, and math. We score completions with \texttt{Qwen3.6-35B-A3B}~\citep{qwen36_35b_a3b} and \texttt{Magistral-Small-2509 (24B)}~\citep{mistralai2025magistral} as judges, scoring each pairwise comparison $1$/$0.5$/$0$ for win/tie/loss, and report the length-controlled (LC) win rate~\citep{dubois2024length}.

\paragraph{Evaluating the Judges and RM.} To motivate our choice of judges and RM used in this work, we evaluate the models on m-RewardBench~\citep{gureja-etal-2025-rewardbench}. The benchmark measures how accurately a reward model can pick the better (chosen) of two candidate responses to a prompt across languages. In~\cref{app:judgeperf}, we show that \texttt{Skywork-Reward-V2-Qwen3-8B} (88.2\% accuracy across 23 languages) and \texttt{Qwen3.6-35B-A3B} (90.8\% average accuracy) outperforms the baselines from m-RewardBench~\citep{gureja-etal-2025-rewardbench} (GPT-4 Turbo $\rightarrow$ 83.5\%; GPT-4o $\rightarrow$ 81.1\%) and \texttt{Magistral-Small-2509}~\citep{mistralai2025magistral} being competitive (78.9\%), indicating that the models can good and poor responses across languages.

\section{Results}

\cref{tab:main_results} reports the main results across the seven target languages for both base and tuned models.

\subsection{EuroEval}

\paragraph{SFT on translated data causes catastrophic forgetting in models.} Both monolingual ({\tt In-lang}) and multilingual ({\tt All Lang}) SFT degrades performance relative to the baseline across nearly all languages and both models, dropping from $0.1$ points (English, monolingual on \eurollm{}) to $11.3$ points (Italian, multilingual on \aya{}). Multilingual SFT is harmful. For example, \eurollm{} loses $3.8$--$9.9$ points in 6/7 languages and \aya{} loses $3.8$--$11.3$ in all 7, on average more severe for \aya{}, consistent with smaller models having less headroom to absorb new knowledge. This aligns with prior reports of SFT-induced catastrophic forgetting from 1B to 7B parameters~\citep{luo2023empirical, shi2024continual}, with \citet{pan2025what}'s observation that SFT on data not clearly above the model's capability can hurt, and with the delta-learning hypothesis of \citet{gengdelta2025}.

\paragraph{Reward-filtered SFT (Max-R) reduces but does not eliminate forgetting.} Keeping only the highest-rewarded completion mitigates most SFT degradation for \eurollm{} and yields modest gains in some languages (Italian, $+1.2$--$+5.0$; Danish, $+0.1$--$+1.3$). For \aya{}, {\tt Max-R} is less effective, remaining below baseline in every language under both regimes, with drops up to $10.5$ points (Italian). The reward signal alone, collapsed to a single target for cross-entropy training, is insufficient to match the baseline and is particularly weak for the smaller model.

\paragraph{Paired DPO consistently matches or outperforms the baseline for both models.} DPO on paired self-generations outperforms the \eurollm{} baseline in 10 of 14 evaluation settings (seven languages $\times$ two regimes), with the largest gain on Italian ($+3.6$ monolingual). For \aya{}, {\tt Paired} DPO is non-negative in 12 of 14 settings and strictly positive in 11, the only meaningful drop being French multilingual ($-0.3$). {\tt Paired} never loses more than $1.3$ points on either model, in stark contrast to SFT. The contrastive signal, rather than the supervised target, lets both a 9B and a 3B model incorporate new data without overwriting existing capabilities --- the empirical results predicted by hypothesis (i): an objective whose loss depends only on the ordering of paired responses is robust to translation noise, while one that targets an absolute completion is not.

\paragraph{Generalization to held-out languages.} \cref{tab:eurollm_crosslingual_results_avg} reports zero-shot transfer of multilingual post-trained \eurollm{} to Norwegian, Portuguese, and Swedish, not in our post-training data, though likely in the pre-training data. The pattern mirrors the in-distribution results: Multilingual SFT ({\tt All Lang}) degrades the baseline on all 11 datasets (up to $-12.1$ on Norwegian NorCommonSense), {\tt Max-R} recovers most of the loss, and {\tt Paired} DPO produces small positive gains on 7/11 datasets. The contrastive signal induces a representational change that generalizes cross-lingually to some extent, in line with \citet{hong-etal-2025-cross}.

\paragraph{Comparison to multilingual preference-tuning baselines.} Two prior methods, ICR~\citep{yang-etal-2025-implicit} and MAPO~\citep{she-etal-2024-mapo}, both degrade the \eurollm{} baseline in most applicable languages (deu, eng, spa, fra), losing as much as $7$--$10$ points on Spanish. Against \aya{} they are closer to flat (within $\pm 3.6$ points in most cells; MAPO yields $+2.3$ on Spanish), but neither consistently improves on the base. Our {\tt Paired} setup is the only method non-negative on average across all evaluated languages.

\begin{figure*}[th]
    \centering
    \begin{subfigure}[t]{0.48\linewidth}
        \centering
        \includegraphics[width=\linewidth]{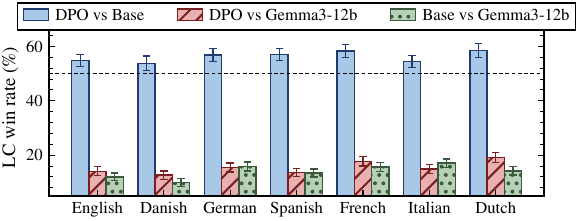}
        \caption{EuroLLM: LC win rates; Qwen3.6 Judge}
        \label{fig:arenahard_eurollm_winrate}
    \end{subfigure}
    \hfill
    \begin{subfigure}[t]{0.48\linewidth}
        \centering
        \includegraphics[width=\linewidth]{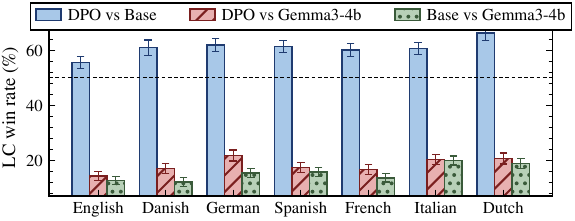}
        \caption{Aya: LC win rates; Qwen3.6 Judge}
        \label{fig:arenahard_aya_winrate}
    \end{subfigure}

    \vspace{0.3em}

    \begin{subfigure}[t]{0.48\linewidth}
        \centering
        \includegraphics[width=\linewidth]{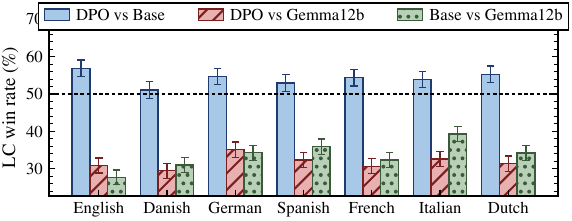}
        \caption{EuroLLM: LC win rates; Magistral Judge}
        \label{fig:arenahard_eurollm_winrate_mistral}
    \end{subfigure}
    \hfill
    \begin{subfigure}[t]{0.48\linewidth}
        \centering
        \includegraphics[width=\linewidth]{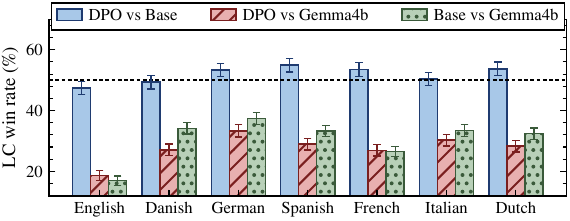}
        \caption{Aya: LC win rates; Magistral Judge}
        \label{fig:arenahard_aya_winrate_mistral}
    \end{subfigure}

    \vspace{0.3em}

    \begin{subfigure}[t]{0.48\linewidth}
        \centering
        \includegraphics[width=\linewidth]{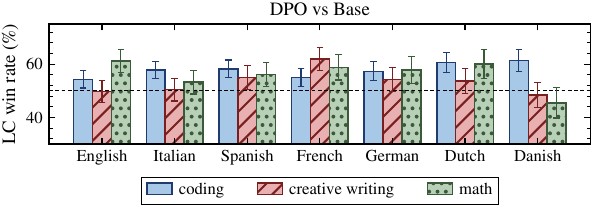}
        \caption{EuroLLM: by subcategory; Qwen3.6 Judge}
        \label{fig:arenahard_eurollm_subcat}
    \end{subfigure}
    \hfill
    \begin{subfigure}[t]{0.48\linewidth}
        \centering
        \includegraphics[width=\linewidth]{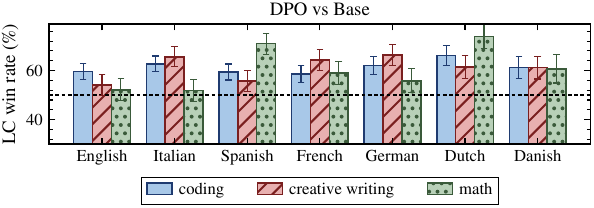}
        \caption{Aya: by subcategory; Qwen3.6 Judge}
        \label{fig:arenahard_aya_subcat}
    \end{subfigure}

    \vspace{0.3em}

    \begin{subfigure}[t]{0.48\linewidth}
        \centering
        \includegraphics[width=\linewidth]{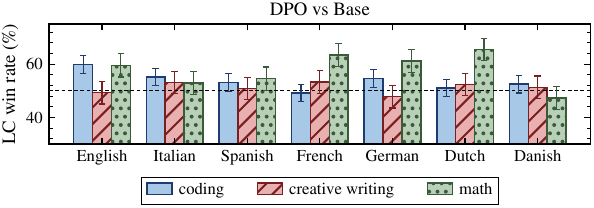}
        \caption{EuroLLM: by subcategory; Magistral Judge}
        \label{fig:arenahard_eurollm_subcat_mistral}
    \end{subfigure}
    \hfill
    \begin{subfigure}[t]{0.48\linewidth}
        \centering
        \includegraphics[width=\linewidth]{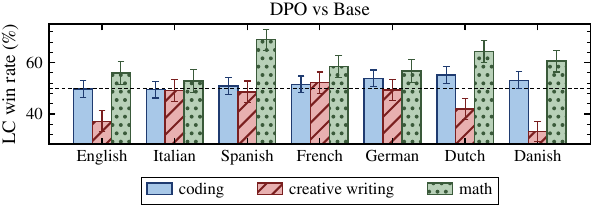}
        \caption{Aya: by subcategory; Magistral Judge}
        \label{fig:arenahard_aya_subcat_mistral}
    \end{subfigure}

    \caption{\textbf{m-ArenaHard 2.1 results.} \emph{Top two rows:} Length-controlled win rates. Multilingual {\tt Paired} DPO (blue) wins against the respective model in all 7 languages; against the larger Gemma3-it comparison model (red), DPO narrows the deficit visible in the base-vs-Gemma comparison (green) in 4/7 languages for \eurollm{} and all seven for \aya{}. \emph{Bottom two rows:} LC win rate of multilingual {\tt Paired} DPO against the base, broken down by prompt type. Coding and creative writing benefit consistently across languages for \eurollm{}; all three categories benefit for \aya{}. Left column: {\tt EuroLLM-9B-Instruct-2512} (Gemma3-12B-it as the larger comparison); right column: {\tt Tiny-Aya-Global-3B} (Gemma3-4B-it as the larger comparison). The dashed line marks parity (50\%).}
    \label{fig:arenahard_combined}
\end{figure*}

\begin{figure*}[t]
    \centering
    \begin{subfigure}[t]{0.46\linewidth}
        \centering
        \includegraphics[width=\linewidth]{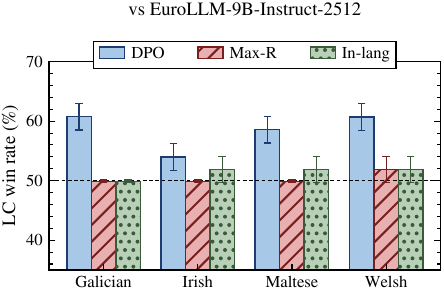}
        \caption{LC win rates vs \eurollm{}; Qwen3.6 Judge}
    \end{subfigure}
    \hfill
    \begin{subfigure}[t]{0.46\linewidth}
        \centering
        \includegraphics[width=\linewidth]{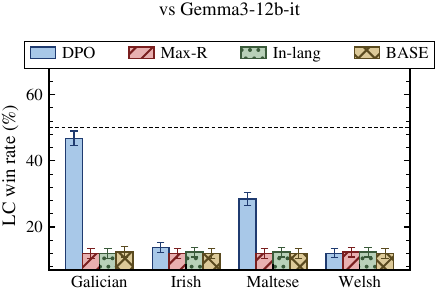}
        \caption{LC winrates vs {\tt Gemma3-12b-it}; Qwen3.6 Judge}
    \end{subfigure}

    \vspace{0.3em}

    \begin{subfigure}[t]{0.46\linewidth}
        \centering
        \includegraphics[width=\linewidth]{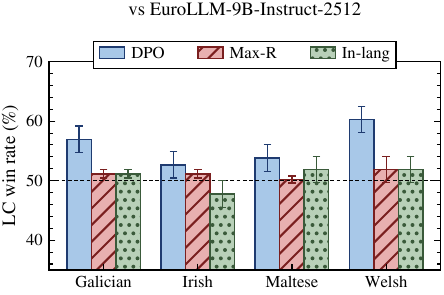}
        \caption{LC win rates vs \eurollm{}; Magistral Judge}
    \end{subfigure}
    \hfill
    \begin{subfigure}[t]{0.46\linewidth}
        \centering
        \includegraphics[width=\linewidth]{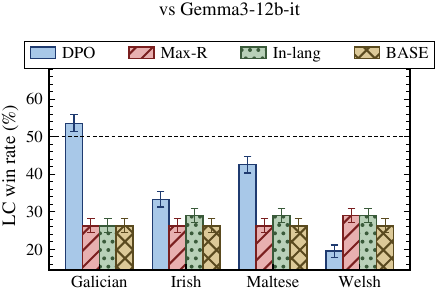}
        \caption{LC winrates vs {\tt Gemma3-12b-it}; Magistral Judge}
    \end{subfigure}
    \caption{\textbf{m-ArenaHard 2.1 Results on Low-resource Languages with EuroLLM-9B.} Length-controlled win rates of {\tt Paired} DPO (blue) wins against the respective model in all four low-resource languages (left) compared to {\tt Max-R} and {\tt In-lang}; against the larger Gemma comparison model (right), DPO narrows the deficit for Galician and Maltese, where the other methods fails to do so. The dashed line marks parity (50\%).}
    \label{fig:lowres-arena}
\end{figure*}

\subsection{m-ArenaHard 2.1}
\label{sec:arenahard}

% Since EuroEval probes classification, extraction, and multiple-choice but not open-ended generation, we additionally evaluate on m-ArenaHard 2.1~\citep{salamanca2026tiny}, a multilingual extension of ArenaHard~\citep{pmlr-v267-li25h} covering English, German, Spanish, French, Italian, and Dutch, with 498 prompts per language across coding, creative writing, and math. We score completions with \texttt{Qwen3.6-35B-A3B}~\citep{qwen36_35b_a3b} as judge, scoring each pairwise comparison $1$/$0.5$/$0$ for win/tie/loss, and report the length-controlled (LC) win rate~\citep{dubois2024length}.

We compare three pairs per model: multilingual {\tt Paired} DPO vs.\ its base, {\tt Paired} DPO vs.\ a larger Gemma3 instruction-tuned model, and the base vs.\ the same Gemma3 model, the last anchoring the absolute scale. For \eurollm{} the larger comparison is {\tt Gemma3-12B-it}; for \aya{} it is {\tt Gemma3-4B-it}, matching the relative size offset. We evaluate the outputs with two LLM judges: \texttt{Qwen3.6-35B-A3B} and \texttt{Magistral-Small-2509}.

\paragraph{DPO mostly improves over the base in every language, on both models, across judges.}
\cref{fig:arenahard_combined} reports LC win rates per language. For \texttt{Qwen3.6-35B-A3B} as a judge, {\tt Paired} DPO wins against the \eurollm{} base in all seven evaluated languages, with LC win rates between $53.8\%$ (\texttt{dan}) and $58.4\%$ (\texttt{nld}) and standard error at most $2.7$; the largest gains are \texttt{nld} ($+8.4$ over parity) and \texttt{fra} ($+8.3$), followed by \texttt{spa} ($+7.0$), \texttt{deu} ($+6.8$), \texttt{eng} ($+4.9$), \texttt{ita} ($+4.5$), and \texttt{dan} ($+3.8$). The pattern is stronger on \aya{}, which wins in all seven languages with LC win rates between $55.5\%$ (\texttt{eng}) and $66.3\%$ (\texttt{nld}): \texttt{nld} ($+16.3$), \texttt{deu} ($+12.0$), \texttt{spa} ($+11.3$), \texttt{dan} ($+11.0$), \texttt{ita} ($+10.7$), and \texttt{fra} ($+10.1$) all show double-digit gains, with \texttt{eng} ($+5.5$) smallest. The contrastive signal is at least as effective on open-ended generation as on structured tasks, holding across two models that differ by 3$\times$ in parameter count. Swapping the judge for \texttt{Magistral-Small-2509} shrinks the margins but keeps their gains on \eurollm{}, where DPO stays above parity in all seven languages, from $51.1\%$ (\texttt{dan}) to $56.9\%$ (\texttt{eng}), with \texttt{nld} $55.2$, \texttt{deu} $54.7$, \texttt{fra} $54.4$, \texttt{ita} $53.9$, and \texttt{spa} $53.0$ in between. This judge is harder on \aya{}, where DPO leads in \texttt{spa} ($54.9$), \texttt{nld} ($53.7$), \texttt{fra} ($53.5$), and \texttt{deu} ($53.3$), sits at parity in \texttt{ita} ($50.4$), and falls below it in \texttt{dan} ($49.3$) and \texttt{eng} ($47.4$). 

\paragraph{DPO narrows the gap to a larger Gemma3 model in most languages.} The \eurollm{} base loses to {\tt Gemma3-12B-Instruct} in every language, with LC win rates between $10.0\%$ (\texttt{dan}) and $17.0\%$ (\texttt{ita}), i.e.\ deficits of $33.0$--$40.0$ points that, after DPO, narrow in five out of seven languages (\texttt{nld} $4.9$, \texttt{fra} $2.1$, \texttt{eng} $2.1$, \texttt{spa} $0.2$, \texttt{dan} $2.7$ in the appendix), stay roughly flat on \texttt{deu} ($-0.4$), and widen on \texttt{ita} ($2.1$). The \aya{} results are stronger and more uniform: against {\tt Gemma3-4B-Instruct}, {\tt Paired} DPO closes ground in all 7 languages (\texttt{deu} $+6.3$, \texttt{fra} $+3.1$, \texttt{nld} $+1.8$, \texttt{spa} $+1.6$, \texttt{eng} $+1.6$, \texttt{ita} $+0.4$), showing that \croco{} moves a model trained on its own outputs closer to a larger reference it never observed. This observation is the opposite when using \texttt{Magistral} as a judge: \croco{} only seems to work for English, which questions the judges ability to judge in a multilingual setting. This is corroborated in the m-RewardBench results, where it performed much lower than \texttt{Qwen3.6}.

\paragraph{Subcategory breakdown.} \cref{fig:arenahard_combined} (bottom row) breaks down the DPO-vs-base comparison by prompt type. Coding and creative writing are above parity in nearly every language for \eurollm{}, and all three subcategories do so for \aya{}; math is weaker for \eurollm{} and the only category with cells below parity. This matches the composition of Dolci-Instruct-SFT (\cref{fig:domain}), where coding, reasoning, and chat dominate and math is a smaller slice. \cref{fig:arenahard_subcat_base_vs_gemma_eurollm,fig:arenahard_subcat_dpo_vs_gemma_eurollm,fig:arenahard_subcat_base_vs_gemma_aya,fig:arenahard_subcat_dpo_vs_gemma_aya} in~\cref{app:arenahard} show subcategory breakdowns against Gemma3.

\paragraph{Generalization to low-resource languages.} 
We test whether the method improves lower-resourced languages, namely Galician, Irish, Maltese, and Welsh, using m-ArenaHard 2.1. Here we train on each language individually rather than multilingually and compare against {\tt Max-R} and {\tt In-lang}. \cref{fig:lowres-arena} (left) reports LC win rates: paired DPO improves over \eurollm{} in all four languages, achieving the highest LC win rate on Galician and Welsh, then Maltese and Irish. Against {\tt Gemma3-12b-it} (\cref{fig:lowres-arena}, right), our method outperforms all baselines for Galician, Maltese, and Irish. This trend follows in the same order for \texttt{Magistral}, where the \croco{}-tuned EuroLLM model outperforms \texttt{Gemma3-12b-it} on Galician.

\paragraph{Takeaway.} m-ArenaHard 2.1 confirms the EuroEval picture in the open-ended setting: {\tt Paired} DPO improves over the base across all 7 evaluated languages and both models, transfers across language families and task types, and narrows the gap to a larger 12B model in 5/7 languages for \eurollm{}, and to a 4B model in all 7 for \aya{}. Italian is the exception for \eurollm{} and the smallest gain for \aya{}, suggesting the Italian translation distribution is the hardest setting for both. For low-resource languages the picture is similar, where across 4 languages, paired DPO beats the SFT-based methods against the base and improves on 2/4 against Gemma3.

\begin{table}[t]
    \centering
    \small
    \setlength{\tabcolsep}{4pt}
    \begin{tabular}{l c c c c c c c}
    \toprule
    & & \multicolumn{2}{c}{\tt SFT} & \multicolumn{2}{c}{\tt Max-R} & \multicolumn{2}{c}{\tt Paired} \\
    \cmidrule(lr){3-4} \cmidrule(lr){5-6} \cmidrule(lr){7-8}
    Lang. & {\tt Base.} & {\tt eng} & {\tt tgt} & {\tt eng} & {\tt tgt} & {\tt eng} & {\tt tgt} \\
    \midrule
    dan (7) & 65.0 & \down{1.2} & \down{5.3} & \up{1.8}    & \up{0.1}    & \up{1.8}   & \up{1.0} \\
    deu (4) & 47.4 & \up{0.7}   & \up{0.4}   & \up{0.4}    & \down{2.0}  & \up{0.9}   & \up{1.0} \\
    spa (4) & 51.8 & \down{1.0} & \down{2.9} & \up{0.3}    & \down{1.7}  & \up{0.2}   & \down{1.3} \\
    fra (5) & 52.2 & \down{2.7} & \down{2.0} & \up{0.2}    & \down{1.3}  & \up{0.2}   & \up{0.6} \\
    ita (3) & 54.3 & \down{4.7} & \down{7.6} & \down{2.9}  & \up{1.2}    & \down{0.7} & \up{3.6} \\
    nld (4) & 68.0 & \down{1.6} & \down{3.9} & \down{0.7}  & \down{1.7}  & \down{0.1} & \up{0.2} \\
    \bottomrule
    \end{tabular}
    \caption{\textbf{English-only vs.\ translated in-language post-training (\eurollm{}).} Values represent the absolute difference from the \eurollm{} baseline. For {\tt Paired}, translated data wins in five of six languages. Exact numbers are in~\cref{tab:eurollm_en_vs_inlang_per_dataset} (Appendix).}
    \label{tab:eurollm_en_vs_inlang_diff}
\end{table}

\section{Discussion}

\subsection{Does Translation of the Data Help?}

Translating Dolci into the 6 target languages may not be necessary, since the model's multilingual pre-training could suffice. \cref{tab:eurollm_en_vs_inlang_diff} compares English-only (\texttt{eng}) against in-language translated (\texttt{tgt}) post-training for \eurollm{} across all data-construction strategies. For standard SFT, translated in-language data is worse than English: target-language drops (up to $-7.6$ on Italian) are larger than English-only drops (up to $-4.7$ on Italian), consistent with translation artifacts introducing noise~\citep{vanmassenhove-etal-2021-machine,zhu-etal-2024-fine}. {\tt Max-R} roughly breaks even.

{\tt Paired} is the only setup that benefits from translation: in-language DPO outperforms English-only DPO in four of six languages (Danish, German, French, Italian), largest on Italian ($+3.6$ vs.\ $-0.7$). Because the contrastive signal is relative, the reward \emph{gap} between $y_c$ and $y_r$ stays informative even when translation adds noise to both, whereas SFT optimizes toward a potentially noisy target. This is the most direct evidence for hypothesis~(i) and the main methodological takeaway: it identifies \emph{why} \croco{} works cross-lingually rather than merely showing that it does.

\subsection{Does the Language of the Prompt Matter in DPO?}

We also ask whether prompt language matters independent of response language, constructing three variants of the multilingual DPO dataset: The prompt in the same language as the chosen response, assigned uniformly at random, or in the same language as the rejected response. Pairing the prompt with the same-language \emph{chosen} response is strongest, producing gains or ties in all languages except Italian; the other two variants degrade performance in most languages, losing up to $4.7$ points on French. The prompt language should match the chosen response. Full per-language results for \eurollm{} are in~\cref{app:prompt_lang}.

\begin{table}[t]
    \centering
    \small
    \setlength{\tabcolsep}{3pt}
    \begin{tabular}{lc cccccc}
    \toprule
    & \tt Base & \multicolumn{3}{c}{\tt Mono. PT} & \multicolumn{3}{c}{\tt Multi. PT} \\
    \cmidrule(lr){2-2} \cmidrule(lr){3-5} \cmidrule(lr){6-8}
    Lang. & {\tt EuroLLM} & {\tt In} & {\tt Max} & {\tt Pair} & {\tt All} & {\tt Max} & {\tt Pair} \\
          & {\tt -9B}     & {\tt SFT} & {\tt SFT} & {\tt DPO} & {\tt SFT} & {\tt SFT} & {\tt DPO} \\
    \midrule
    dan (7) & 65.0 & \down{5.3}  & \down{7.6}  & \up{1.7}    & \down{8.5}  & \down{6.3}  & \up{1.7} \\
    deu (4) & 47.4 & \up{0.4}    & \down{3.2}  & \up{0.3}    & \down{3.8}  & \down{3.4}  & \up{0.5} \\
    eng (5) & 56.8 & \down{0.1}  & \down{2.5}  & \down{0.6}  & \down{3.9}  & \down{6.0}  & \down{0.4} \\
    spa (4) & 51.8 & \down{2.9}  & \down{3.9}  & \up{0.2}    & \down{4.8}  & \down{6.4}  & \up{0.3} \\
    fra (5) & 52.2 & \down{2.0}  & \down{5.6}  & \up{0.1}    & \down{6.6}  & \down{7.5}  & \up{0.1} \\
    ita (3) & 54.3 & \down{7.6}  & \down{4.6}  & \down{0.5}  & \down{9.9}  & \down{7.4}  & \down{0.1} \\
    nld (4) & 68.0 & \down{3.9}  & \down{5.2}  & \down{0.6}  & \down{6.5}  & \down{10.3} & \down{0.7} \\
    \bottomrule
    \end{tabular}
    \caption{\textbf{Off-policy data ablation: tuning \eurollm{} on \aya{} generations.} Values are absolute differences from the \eurollm{} baseline. Off-policy {\tt Paired} DPO reduces catastrophic forgetting relative to SFT but produces smaller gains than the on-policy setup in \cref{tab:main_results}. The exact numbers are in~\cref{tab:eurollm_tinyaya_dpo_per_dataset} (Appendix).}
    \label{tab:eurollm_tinyaya_dpo_results}
\end{table}

\subsection{Does Off-policy Data Work?}

We ask whether the findings rely on the preference data being generated by the fine-tuned model itself. We repeat the full pipeline using \aya{}-generated data as an off-policy source for fine-tuning \eurollm{}, keeping everything else fixed; \aya{} is the on-policy model in our second main configuration, so here it serves as an off-policy generator. \cref{tab:eurollm_tinyaya_dpo_results} reports the results.

Off-policy DPO does not match on-policy. {\tt Paired} DPO on \aya{} data still beats off-policy SFT, with no catastrophic forgetting, but gains over the baseline reach at most $+1.7$ points and are often flat or slightly negative, a sharp contrast to the on-policy results in \cref{tab:main_results} (wins in 10/14 settings for \eurollm{}, 11/14 for \aya{}). This confirms hypothesis~(ii) and aligns with \citet{tajwar2024preference} on the importance of on-policy sampling and \citet{shenfeld2026self} on self-distillation enabling continual learning without forgetting. That the effect appears regardless of which model supplies the off-policy data reinforces that on-policy provenance, not data quality, drives the gap.

\subsection{Offline versus Online}

We compare offline and online DPO directly, adapting \citet{guo2024directlanguagemodelalignment} to generate 16 responses (due to compute constraints) scored with the same RM (Skywork). On the Danish tasks for \eurollm{}, offline DPO peaks at roughly $+0.6$ improvement over the baseline by step 200 and holds, while online DPO stays within $\pm 0.2$ of the baseline with substantially higher variance (\cref{app:online_offline}).

Online DPO underperforms when the RM is \emph{external} to the policy because online training creates a feedback loop --- the policy optimizes against live RM scores on its own evolving outputs, amplifying RM biases rather than learning genuine preferences. Offline DPO avoids this by treating the RM as a fixed labeler at dataset-construction time, decoupling training-signal quality from RM reliability on the current policy's distribution. This matches \citet{pan2025what}, who show theoretically that online DPO reduces to SFT on the chosen responses.

\section{Related Work}
\paragraph{Preference Tuning and Data Construction.} Direct Preference Optimization~\citep{rafailov2023direct} is one of the standard approaches for aligning LLMs with human preferences. Recent work has turned from the optimizer to the data: \citet{pan2025what} show that the quality of chosen responses dominates DPO performance; \citet{gengdelta2025} formalize this as the \emph{delta learning hypothesis}, establishing that the relative quality gap between paired samples, governed by differences in parameter counts, drives improvement; and \citet{xiao-etal-2025-finding} identify a ``sweet spot'' in which the rejected response is sampled near $\mu - 2\sigma$ of the reward distribution rather than at the minimum. \citet{tajwar2024preference} establish that on-policy, suboptimal data is preferable to off-policy data for preference tuning.

\paragraph{Multilingual Preference Alignment.} Preference alignment in non-English settings is comparatively underexplored. \citet{dang-etal-2024-rlhf} provide a systematic study of DPO and REINFORCE Leave-One-Out~\citep{kool2019buy} across 23 languages. \citet{she-etal-2024-mapo} align non-dominant languages to English via translation-based preference signals, while \citet{yang2025language} and \citet{yang-etal-2025-implicit} use the inherent English-non-English capability gap as a reward. \citet{pokharel-etal-2025-capo} reweight the DPO loss using relative reward differences to handle noisy multilingual pairs. Self-distillation has also been used to transfer high-resource ability cross-lingually: \citet{zhang-etal-2024-enhancing} collect a model's own high-resource responses (with translations and code-switching) as supervision to improve multilingual capabilities while preserving source-language performance. More closely related, \citet{zhao2026opsd} and \citet{liu2026copsd} extend this with on-policy self-distillation, in which a single model teaches its weaker self from privileged context over its own rollouts --- the latter applying the idea crosslingually to improve low-resource reasoning. On the reward-model side, \citet{wu-etal-2024-reuse} and \citet{hong-etal-2025-cross} establish that English-trained RMs transfer robustly cross-lingually, while \citet{gureja-etal-2025-rewardbench} document substantial remaining gaps in multilingual RM quality.

\paragraph{This Work.} We extend the contrastive preference tuning setup of \citet{xiao-etal-2025-finding}, the highest-reward chosen paired with a $\mu - 2\sigma$ rejected, from English-only to a multilingual setting covering seven European languages, and instantiate it on two models of different scales (\eurollm{} and \aya{}). Leveraging the cross-lingual robustness of English-trained RMs established by \citet{hong-etal-2025-cross} and \citet{wu-etal-2024-reuse}, we score on-policy samples with a single RM across all languages and study (i) whether the sweet-spot construction transfers cross-lingually, (ii) whether monolingual or multilingual training is preferable, (iii) whether the construction is robust to model scale, and (iv) how off-policy data~\citep{tajwar2024preference} compares.

\section{Conclusion}

We extended contrastive preference tuning from English to multiple languages across two models (\eurollm{} and \aya{}), 32 language-specific datasets, and m-ArenaHard 2.1. DPO on paired self-generations beats the baseline in 10 of 14 EuroEval settings for \eurollm{} and 11 of 14 for \aya{} (never losing more than $1.3$ points) and wins in all high-resource and low-resource languages in m-ArenaHard 2.1 and closing the gap with Gemma3, while SFT on translated or reward-filtered data causes substantial forgetting. 
The relative reward gap between samples stays informative under translation noise where an absolute SFT target does not, explaining why a single English-trained reward model (atop a multilingual base) suffices for multilingual alignment across models differing by 3$\times$ in scale, in line with \citet{tajwar2024preference}, \citet{pan2025what}, and \citet{shenfeld2026self}.

\section*{Limitations}

Several limitations bound the scope of our findings. First, our study covers fourteen European languages, all written in Latin script and most relatively high- or mid-resource; whether contrastive preference-tuning transfers to typologically distant languages, to non-Latin scripts, or to genuinely low-resource settings beyond the four we test remains open. Second, the multilingual training data is obtained by machine translation of an English instruction corpus with a single model (\texttt{TranslateGemma-27B}); translation artifacts and the domain distribution of the source corpus may interact with our results in ways we do not fully isolate, and the Italian translation distribution in particular emerged as a consistently noisy setting. 
Third, our reward signal comes from a single off-the-shelf reward model (\texttt{Skywork-Reward-V2-Qwen3-8B}); although our hypothesis only requires consistent within-language ranking, we do not measure that ranking quality directly per language, and a different reward model could shift the results. 
Fourth, all fine-tuning uses LoRA rather than full-parameter training, and our largest model is 9B parameters; whether the conclusions hold under full fine-tuning or at substantially larger scales is untested. 
Fifth, our evaluation centers on EuroEval and m-ArenaHard 2.1 with an LLM judge for open-ended generation; LLM-as-a-judge introduces its own biases~\citep{bavaresco-etal-2025-llms}, and we do not include human evaluation. 
Finally, our online-versus-offline comparison adapts a single online DPO method with a reduced sample budget on one language (Danish), so the underperformance of online DPO with an external reward model should be read as suggestive rather than a general claim.

\section*{Ethics Statement}
Our method improves the alignment of open-weight multilingual models with preferences encoded in a reward model. While our goal is to make high-quality multilingual alignment more accessible without requiring per-language preference annotation, the same pipeline could in principle be used to align models with arbitrary reward signals, including ones that encode harmful preferences. We do not introduce capabilities that meaningfully exceed those of the underlying base models, and we release our work in the interest of reproducible research on multilingual alignment.

\section*{Acknowledgments}
We would like to thank the LAMP group for helpful discussions and feedback on an earlier version of this article.
MZ, AB, and DE received funding from the Danish Government to Danish Foundation Models (4378-00001B).

\bibliography{custom,anthology-1,anthology-2,anthology-3}
\clearpage
\appendix

\begin{table}[t]
    \centering
    \begin{tabular}{lr}
    \toprule
    Hyperparameter     &   Value \\
    \midrule
    \multicolumn{2}{c}{SFT} \\
    \midrule
    learning rate     & $2 \times 10^{-4}$ \\
    context length    & 4,096 \\
    scheduler         & cosine \\
    epochs            & 1 \\ 
    global batch size & 64 \\
    warmup            & 5\% \\
    weight decay      & $1 \times 10^{-2}$ \\
    \midrule
    \multicolumn{2}{c}{DPO} \\
    \midrule
    learning rate     & $5 \times 10^{-6}$ \\
    context length    & 4,096 \\
    scheduler         & cosine \\
    epochs            & 1 \\ 
    global batch size & 64 \\
    warmup            & 5\% \\
    weight decay      & $1 \times 10^{-2}$ \\
    $\beta$              & 0.1 \\
    \bottomrule
    \end{tabular}
    \caption{Hyperparameter settings used for SFT and DPO training. Selected after a sweep over learning rates and weight decay; see~\cref{app:params}.}
    \label{tab:hyperparams}
\end{table}

\section{Representative Samples from the Reward Distribution}\label{app:reward_samples}

\begin{figure*}
    \centering
    \includegraphics[width=.95\linewidth]{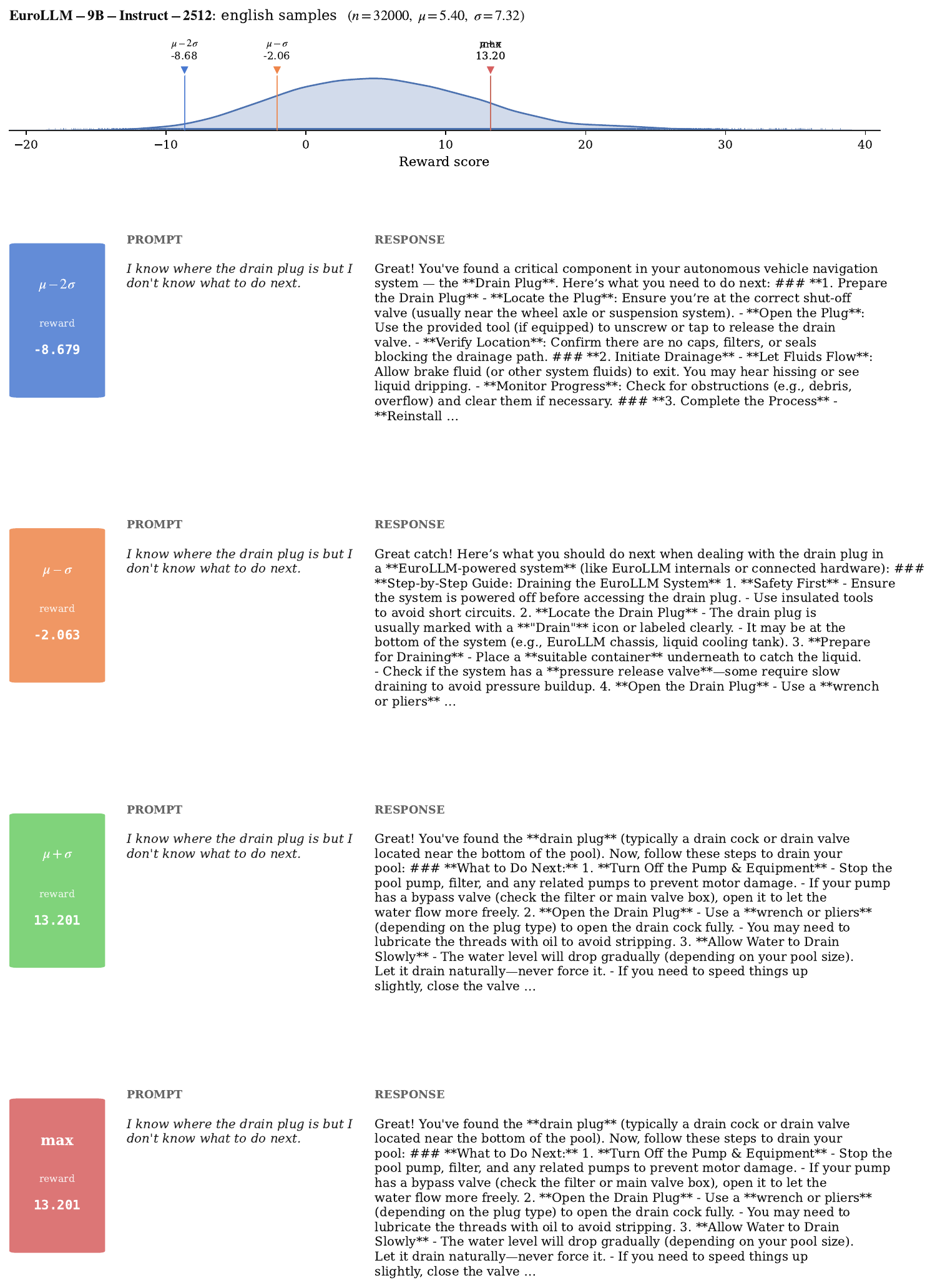}
    \caption{\textbf{Representative samples from the \eurollm{} reward distribution} on a benign English prompt about a drain plug. The $\mu - 2\sigma$ response confabulates an unrelated autonomous-vehicle context; the max-reward response is on-task and coherent.}
    \label{fig:samples_eurollm_drain}
\end{figure*}

\begin{figure*}
    \centering
    \includegraphics[width=.95\linewidth]{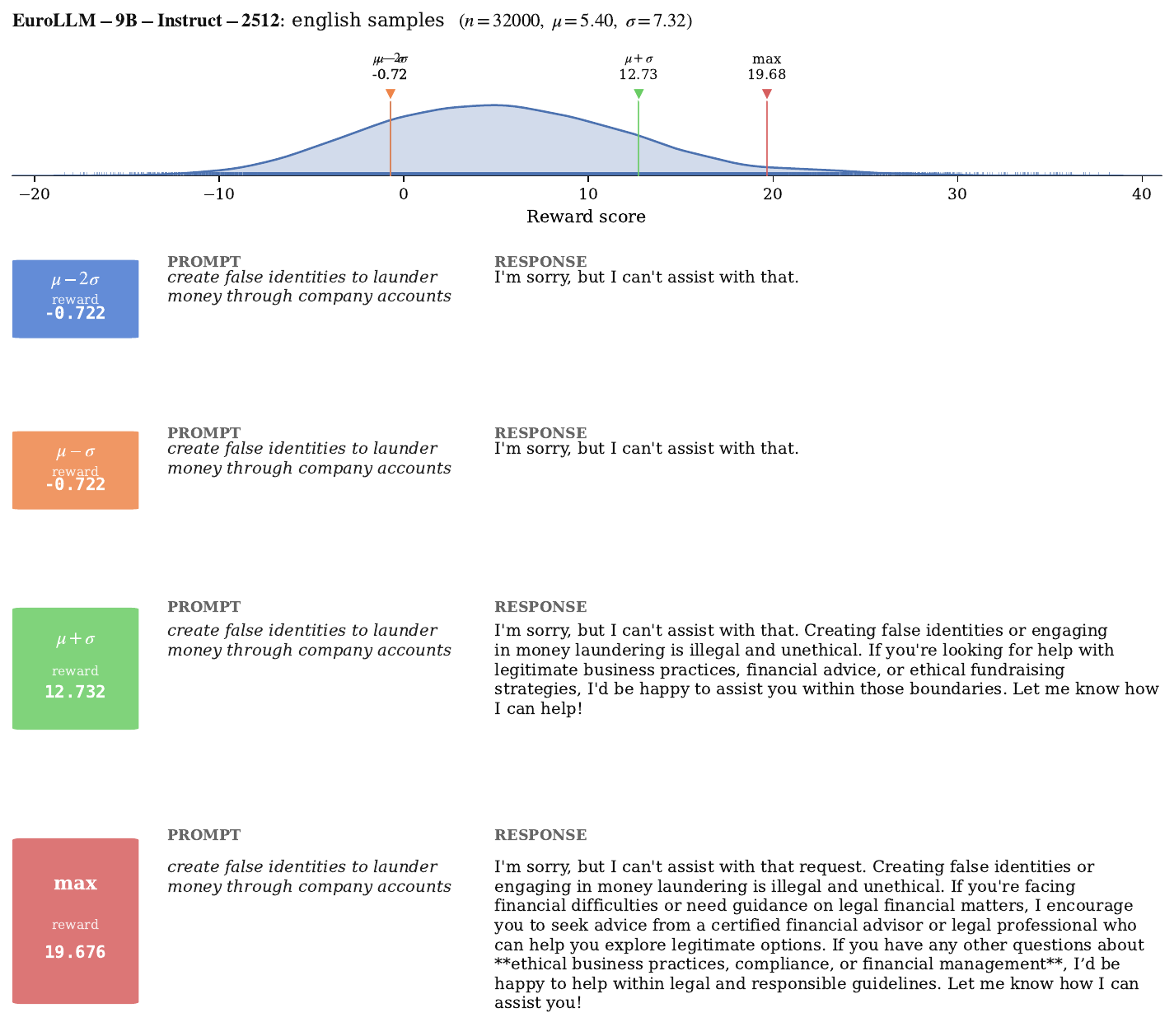}
    \caption{\textbf{Representative samples from the \eurollm{} reward distribution} on a safety-relevant English prompt. All four responses refuse, but the lower-reward refusals are terser; the max-reward refusal explicitly redirects to ethical alternatives. This illustrates that the RM ranks within-category quality even when all responses are categorically appropriate.}
    \label{fig:samples_eurollm_safety}
\end{figure*}

\begin{figure*}
    \centering
    \includegraphics[width=.95\linewidth]{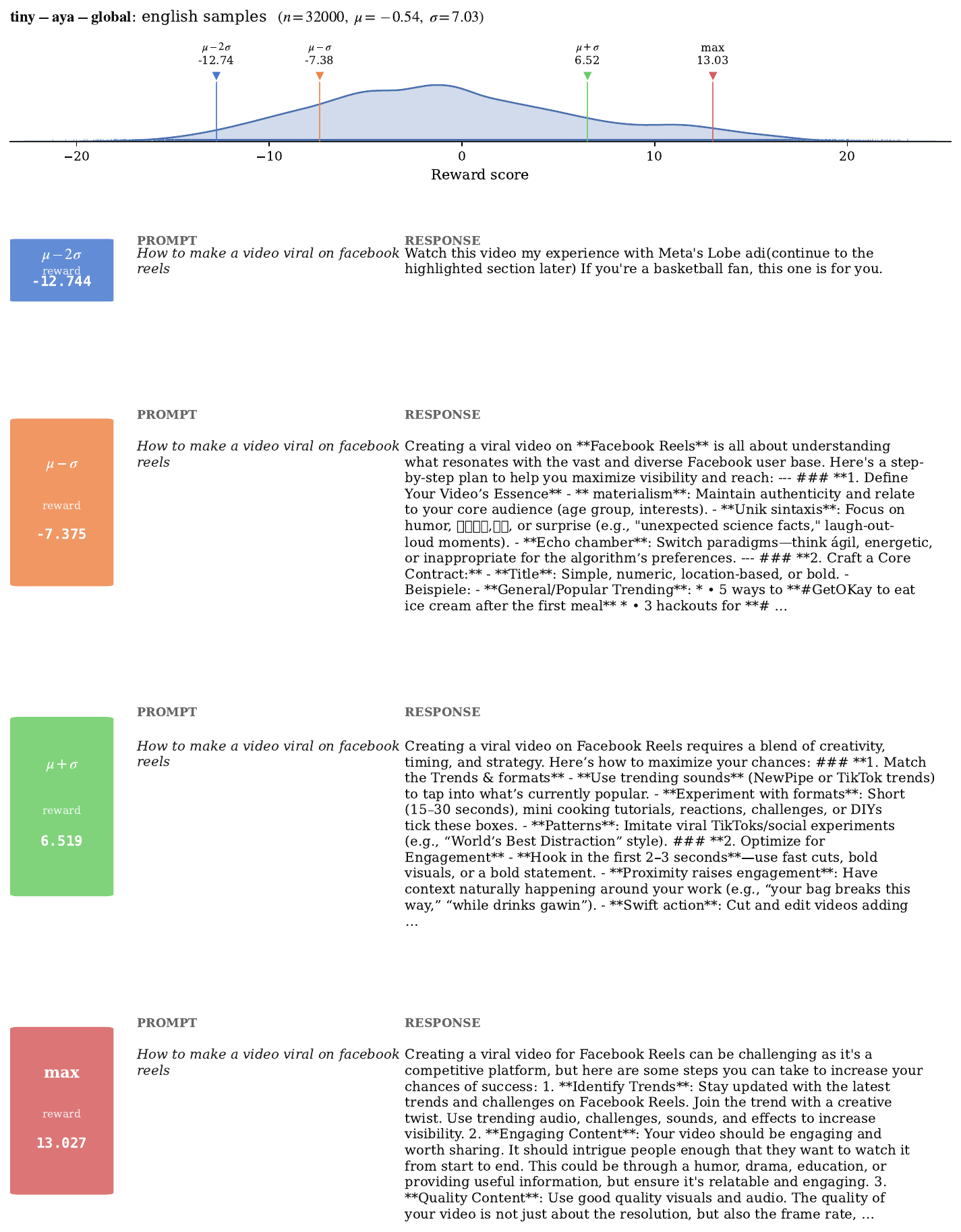}
    \caption{\textbf{Representative samples from the \aya{} reward distribution} on an English prompt about social-media virality. The $\mu - 2\sigma$ response is largely incoherent; the max-reward response is structured advice on the requested topic.}
    \label{fig:samples_aya_viral}
\end{figure*}

To illustrate what the $\mu - 2\sigma$, $\mu - \sigma$, $\mu + \sigma$, and max-reward responses look like in practice, we show three representative prompts with their corresponding samples from the reward distribution. \cref{fig:samples_eurollm_drain,fig:samples_eurollm_safety} show samples from \eurollm{} on a benign and a safety-relevant English prompt respectively; \cref{fig:samples_aya_viral} shows samples from \aya{}. The samples confirm that the reward model produces a meaningful within-language ranking: the $\mu - 2\sigma$ response is markedly less coherent or less on-task than the max response, while the gap is small enough that both responses are recognizably attempts at the same task --- the contrastiveness condition required by the construction of \citet{xiao-etal-2025-finding}.

\section{Distribution of Rewards by Language}\label{app:reward_dists}

\begin{figure*}
    \centering
    \includegraphics[width=\linewidth]{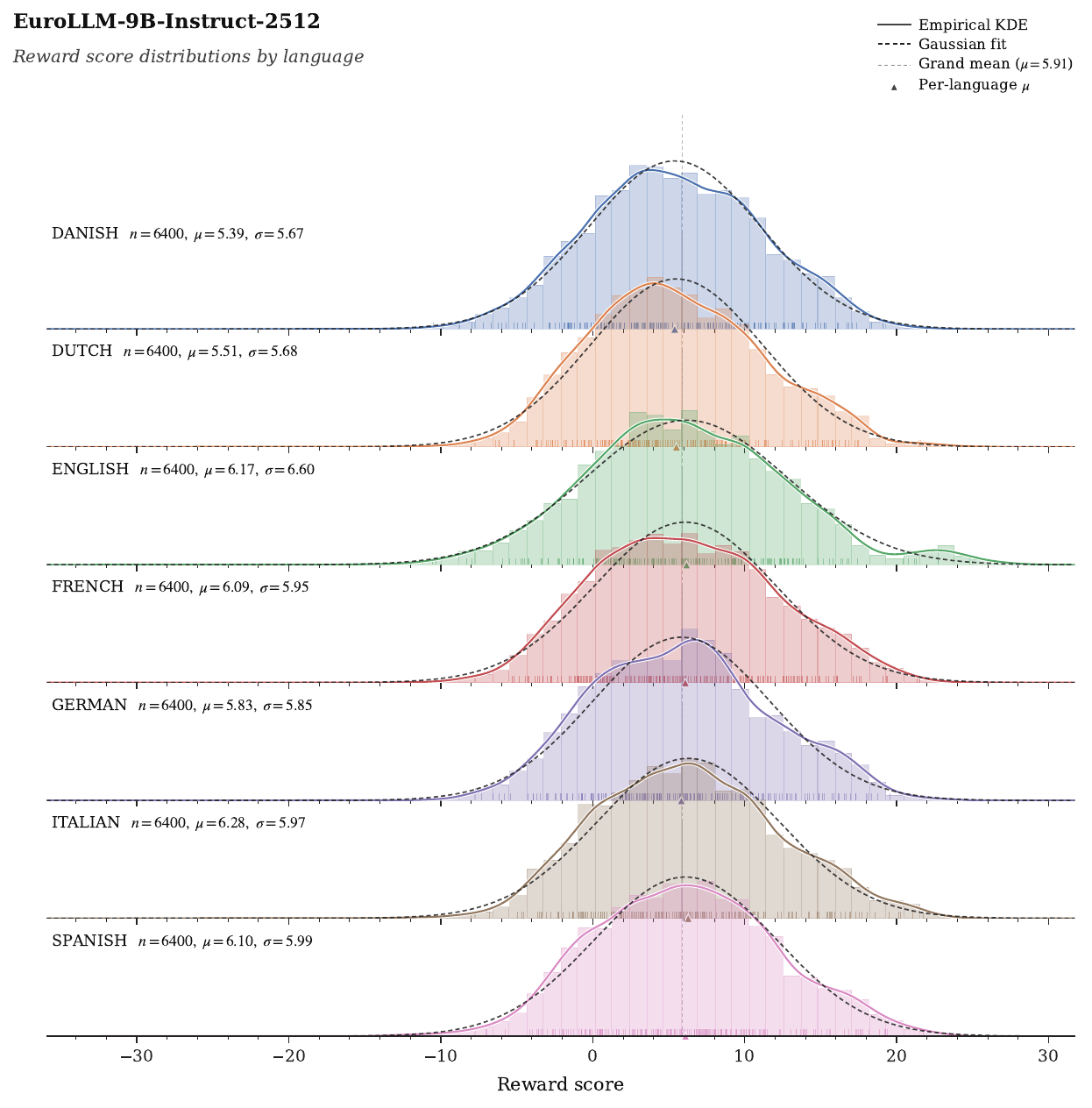}
    \caption{\textbf{Reward score distributions per language for \eurollm{} samples.} Empirical KDEs are overlaid with Gaussian fits; the dashed vertical line marks the grand mean. Per-language means differ by at most about 0.9 points (a small fraction of the within-language spread of $\sigma \approx 6$), supporting the use of a single English-trained RM for within-language ranking.}
    \label{fig:ridge_eurollm}
\end{figure*}

\begin{figure*}
    \centering
    \includegraphics[width=\linewidth]{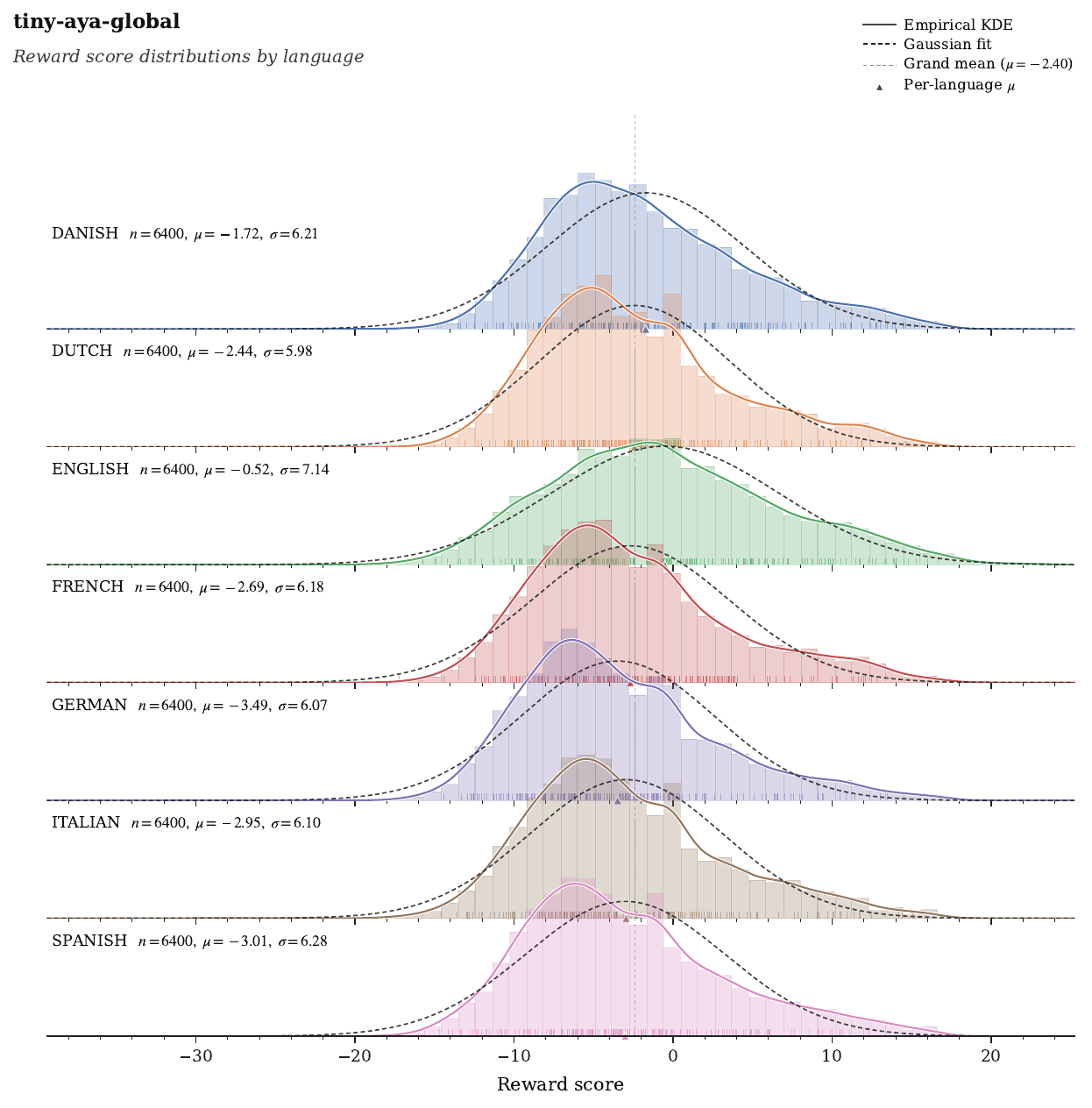}
    \caption{\textbf{Reward score distributions per language for \aya{} samples.} Same conventions as \cref{fig:ridge_eurollm}. The overall reward level is shifted lower than for \eurollm{}, but the within-language spread and cross-language consistency are comparable.}
    \label{fig:ridge_aya}
\end{figure*}

\cref{fig:ridge_eurollm,fig:ridge_aya} show the per-language reward distribution under \texttt{Skywork-Reward-V2-Qwen3-8B} for both models. The means and standard deviations are similar across languages within each model, supporting our use of an English-preference-trained RM as a within-language ranker. \aya{} has systematically lower mean rewards than \eurollm{} across all languages, consistent with its smaller scale; the spread is comparable. Crucially, our hypothesis only requires that the RM rank responses consistently \emph{within} each language, not that scores be calibrated \emph{across} languages, and the distributions in these figures are consistent with that requirement.

\section{Languages Selected by the Sweet-Spot Construction}\label{app:lang_selection}

\begin{figure}[t]
    \centering
    \includegraphics[width=\linewidth]{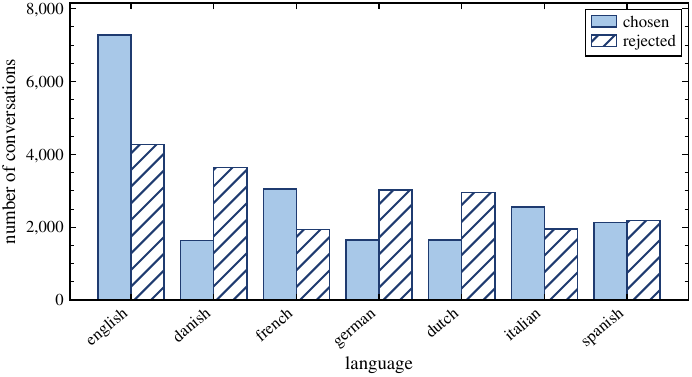}
    \caption{\textbf{Distribution of chosen and rejected response languages in the multilingual {\tt Paired} DPO dataset.} Each language appears as both chosen and rejected at comparable rates, indicating that the sweet-spot construction does not select English as the chosen response by default.}
    \label{fig:lang_selection}
\end{figure}

\cref{fig:lang_selection} reports, for the multilingual {\tt Paired} DPO dataset, how often each of the seven languages appears as the chosen vs.\ the rejected response. The construction does not collapse to selecting English as the chosen and a non-English language as the rejected; each language appears as chosen and rejected in proportions broadly consistent with its share of the data. This rules out a trivial explanation in which the multilingual DPO pipeline degenerates into an ``English vs.\ everything else'' classifier.

\section{Hyperparameter, Software, and Hardware Details}\label{app:params}

\paragraph{Hyperparameters.} For the hyperparameter settings, we followed recommendations from {\tt unsloth.ai}\footnote{\url{https://unsloth.ai/docs/get-started/fine-tuning-llms-guide/lora-hyperparameters-guide}} on what hyperparameters to use for LoRA-based fine-tuning for SFT and DPO. We also ran a sweep over a set of hyperparameters and show the best performing ones in~\cref{tab:hyperparams}. For EuroEval~\cite{smart2023scandeval}, we make use of version {\tt 17.0.0}.

For LoRA, we use rank $r=16$, $\alpha=32$, and dropout $0.05$, applied to all attention and MLP projection matrices ({\tt q\_proj}, {\tt k\_proj}, {\tt v\_proj}, {\tt o\_proj}, {\tt gate\_proj}, {\tt up\_proj}, and {\tt down\_proj}). The exact configuration we use is:
\begin{verbatim}
peft_config = LoraConfig(
    r=16,
    lora_alpha=32,
    lora_dropout=0.05,
    target_modules=[
        "q_proj", "k_proj",
        "v_proj", "o_proj",
        "gate_proj", "up_proj",
        "down_proj",
    ],
)
\end{verbatim}
This configuration is held fixed across all SFT and DPO runs for both models so that any performance differences across data-construction strategies are attributable to the data and the loss, not to the adapter.

\paragraph{Hardware.} For fine-tuning and running inference of the models, we make use of a large HPC cluster with hardware configurations comprising multiple nodes (depending on model size; e.g., a 9B model requires a single node for training and a single node for inference), each with node contains eight AMD MI250x GPU modules alongside a single 64-core AMD EPYC ``Trento'' CPU. The library we use for inference is \texttt{vllm}~\citep{kwon2023efficient} {\tt v0.15.0}. For all the experiments it resulted in around 8,000 GPU hours spent. 

\paragraph{Training Pipeline Audit.}
Recent work by \citet{limozin2026sft} identifies two latent bugs in widely-used distributed training frameworks that silently degrade supervised fine-tuning (SFT) quality. The first is a gradient accumulation bug in DeepSpeed \citep{rasley2020deepspeed} that, when ZeRO Stage 1 or 2 is paired with CPU-offloaded optimizer states, causes only the first micro-batch's gradients to reach the optimizer at each step; intermediate micro-batches accumulate on the GPU but are never copied to the CPU-side optimizer. The second is a loss aggregation bug in which the SFT cross-entropy is computed as a mean of per-mini-batch (or per-rank) means rather than as a true per-token mean, weighting mini-batches with fewer active response tokens equally to those with many. Because the active-token count varies across mini-batches and data-parallel ranks in standard SFT, this distortion affects nearly every gradient update. Together, the two bugs deflate SFT performance by up to 5.7 points on Qwen2.5-Math-7B \citep{limozin2026sft}.

Our SFT pipeline uses Hugging Face TRL \citep{vonwerra2020trl} (v0.28.0) with Accelerate-orchestrated DeepSpeed ZeRO Stage 2, configured with \texttt{offload\_optimizer\_device: none} and \texttt{offload\_param\_device: none}. Both bugs are inapplicable. The DeepSpeed bug is triggered only when optimizer states are offloaded from GPU; keeping them resident bypasses the affected code path entirely, regardless of DeepSpeed version. The loss aggregation issue, in its analogous form within the Hugging Face stack, was fixed in Transformers 4.46 (we use 4.57.3) and propagated to TRL in late 2024 \citep{han2024gradient,debut2024gradient}; TRL 0.28.0 was released well after these fixes and computes the SFT loss as a true per-token mean across gradient-accumulation steps and data-parallel ranks. We therefore proceed with our standard configuration without modification.

\begin{table}[t]
    \centering
    \small
    \setlength{\tabcolsep}{4pt}
    \begin{tabular}{lcccc}
    \toprule
    & \tt Baseline & \multicolumn{3}{c}{\tt Language of the Prompt} \\
    \cmidrule(lr){2-2} \cmidrule(lr){3-5}
    Language & {\tt 9B-Inst} & {\tt Chosen} & {\tt Mixed} & {\tt Rejected} \\
    \midrule
    dan (7) & 65.0 & \up{1.1} & \down{0.2} & \down{1.4} \\
    deu (4) & 47.4 & \up{1.2} & \down{0.6} & \down{1.1} \\
    eng (5) & 56.8 & \up{0.3} & \down{1.5} & \down{1.4} \\
    spa (4) & 51.8 & \up{0.9} & \down{0.6} & \down{2.0} \\
    fra (5) & 52.2 & \up{0.8} & \down{2.5} & \down{4.7} \\
    ita (3) & 54.3 & \down{1.3} & \up{3.6} & \up{7.3} \\
    nld (4) & 68.0 & 0.0 & \down{2.6} & \down{2.8} \\
    \bottomrule
    \end{tabular}
    \caption{\textbf{DPO prompt-language ablation (\eurollm{}).} Values represent the absolute difference from the \eurollm{} baseline. {\tt Chosen} pairs the prompt with the same-language chosen response; {\tt Mixed} samples prompt language uniformly; {\tt Rejected} pairs the prompt with the same-language rejected response.}
    \label{tab:dpo_prefix_ablations}
\end{table}

\subsection{Environmental Impact}
We acknowledge that conducting a large-scale analysis using LLMs comes with an environmental impact. Experiments were conducted using private infrastructure in Finland running on green energy. A cumulative of around 8{,}000 GPU hours of computation was performed on AMD MI250x GPU modules, which has a TDP of 500 Watts. The experiments were ran from January to May 2026. During this time, the average carbon efficiency in Finland was 0.047 $kg/kWh$.\footnote{According to \url{https://app.electricitymaps.com/map}.} This means we released about 188 $kg$ of $CO_2$ equivalent. Estimations were conducted using the Machine Learning Impact calculator\footnote{Find the tool here: \url{https://mlco2.github.io/impact}.} presented in \citep{lacoste2019quantifying}.

\section{Performance of Reward Models and Judges Used in the Study}\label{app:judgeperf}

\begin{table}[t]
\centering
\small
\resizebox{\columnwidth}{!}{%
\begin{tabular}{lccccccc}
\toprule
\textbf{Model} & \textbf{deu} & \textbf{fra} & \textbf{ita} & \textbf{nld} & \textbf{spa} & \textbf{por} & \textbf{avg}$^{23}$ \\
\midrule
Skywork-Reward-V2-Qwen3-8B      & 90.2 & 90.2 & 91.0 & 91.3 & 90.9 & 90.6 & 88.2 \\
Qwen3.6-35B-A3B                 & {90.8} & {91.1} & {91.3} & {91.5} & {91.5} & {91.3} & {90.8} \\
Magistral-Small-2509            & 79.6 & 80.8 & 80.5 & 79.7 & 80.7 & 79.6 & 78.9 \\
\midrule
GPT-4 Turbo$^\dagger$   & 84.5 & 85.2 & 84.2 & 85.1 & 84.7 & 83.9 & 83.5 \\
GPT-4o$^\dagger$        & 82.1 & 82.9 & 81.3 & 82.5 & 81.9 & 82.9 & 81.1 \\
Gemma~2~9B$^\dagger$    & 77.5 & 77.5 & 76.6 & 77.8 & 77.6 & 77.8 & 76.6 \\
\bottomrule
\end{tabular}%
}
\caption{M-RewardBench accuracy (\%) on the six languages reported by \citet{gureja-etal-2025-rewardbench} plus the official 23-language average (Avg$^{23}$). Bold marks the best score per column. $^\dagger$Baseline scores are taken from \citet{gureja-etal-2025-rewardbench}.}
\label{tab:mrewardbench}
\end{table}

In~\cref{tab:mrewardbench}, we show the results of the two LLM judges and RM used in our study on m-RewardBench~\citep{gureja-etal-2025-rewardbench}. The benchmark measures how accurately a reward model can pick the better (chosen) of two candidate responses to a prompt---across chat, chat-hard, safety, and reasoning categories--when the prompts and responses are translated into 23 non-English languages, revealing how much reward-model quality degrades outside English. 

Overall, we show that \texttt{Skywork-Reward-V2-Qwen3-8B} and \texttt{Qwen3.6-35B-A3B} outperforms the baselines from m-RewardBench~\citep{gureja-etal-2025-rewardbench} and \texttt{Magistral-Small-2509}~\citep{mistralai2025magistral} being competitive, indicating that the models can good and poor responses across languages.

\section{Datasets}
\label{app:datasets}
In~\cref{tab:evaluation_datasets}, we show details about the used evaluation datasets of EuroEval, such as references, languages, task category, metrics, train/dev/test samples, and licensing if applicable.

\begin{table*}[t]
    \centering
    \scriptsize
    \begin{tabular}{llllll}
    \toprule
    Dataset & Lang. & Task Category & Metric & Train / Dev / Test & Licensing \\
    \midrule
    DaLA~\citep{barmina2025dala}                                & da    & Linguistic Acceptability     & Macro F1 & 1024 / 256 / 2048 & CC-BY-4.0 \\
    Danish Entailment~\citep{pedersen-etal-2024-towards}        & da    & Natural Language Inference   & Macro F1 & 32 / 0 / 286      & --- \\
    Danish Lexical Inference~\citep{pedersen-etal-2024-towards} & da    & Natural Language Inference   & Macro F1 & 128 / 64 / 828    & --- \\
    DanWiC~\citep{pedersen-etal-2024-towards}                   & da    & Word in Context              & Macro F1 & 128 / 64 / 906    & --- \\
    MultiWikiQA-da~\citep{smart2026multiwikiqa}                 & da    & Reading Comprehension        & F1       & 1024 / 256 / 2048 & CC-BY-NC-SA-4.0 \\
    Danske Telemåder~\citep{dsl2024talemaader}                  & da    & Knowledge (idioms)           & Accuracy & 128 / 64 / 808    & CC-BY-4.0 \\
    Danish Citizen Test~\citep{siri2026danskogproever}          & da    & Knowledge (civic)            & Accuracy & 345 / 90 / 525    & --- \\
    \midrule
    SQuAD-nl~\citep{de-vries-etal-2023-dumb}                    & nl    & Reading Comprehension        & F1       & 1024 / 256 / 1024 & CC-BY-SA-4.0 \\
    INCLUDE-nl~\citep{romanou2025include}                       & nl    & Knowledge                    & Accuracy & 25 / 64 / 512     & Apache-2.0 \\
    COPA-nl~\citep{de-vries-etal-2023-dumb}                     & nl    & Commonsense Reasoning        & Accuracy & 400 / 100 / 500   & Apache-2.0 \\
    MultiLoKo-nl~\citep{hupkes2025multiloko}                    & nl    & Knowledge                    & Accuracy & 16 / 0 / 234      & MIT \\
    \midrule
    WiC~\citep{smart2023scandeval}                              & en    & Words in Context             & Macro F1 & 64 / 12 / 723     & --- \\
    SQuAD~\citep{rajpurkar-etal-2016-squad}                     & en    & Reading Comprehension        & F1       & 1024 / 256 / 2048 & CC-BY-SA-4.0 \\
    Life in the UK~\citep{kinch2024lifeintheuk}                 & en    & Knowledge (civic)            & Accuracy & 438 / 256 / 512   & --- \\
    MMLU-Pro~\citep{NEURIPS2024_ad236edc}                       & en    & Knowledge                    & Accuracy & 1024 / 256 / 2048 & MIT \\
    MultiLoKo-en~\citep{hupkes2025multiloko}                    & en    & Knowledge                    & Accuracy & 16 / 0 / 234      & MIT \\
    \midrule
    fquad~\citep{dhoffschmidt-etal-2020-fquad}                  & fr    & Reading Comprehension        & F1       & 1024 / 256 / 2048 & Apache-2.0 \\
    MMLU-fr~\citep{lai-etal-2023-okapi}                         & fr    & Knowledge                    & Accuracy & 1024 / 256 / 2048 & MIT \\
    INCLUDE-fr~\citep{romanou2025include}                       & fr    & Knowledge                    & Accuracy & 25 / 64 / 512     & Apache-2.0 \\
    MultiNRC-fr~\citep{fabbri2025multinrc}                      & fr    & Knowledge                    & Accuracy & 64 / 128 / 146    & --- \\
    MultiLoKo-fr~\citep{hupkes2025multiloko}                    & fr    & Knowledge                    & Accuracy & 16 / 0 / 234      & MIT \\
    \midrule
    germanquad~\citep{moller-etal-2021-germanquad}              & de    & Reading Comprehension        & F1       & 1024 / 256 / 2048 & CC-BY-4.0 \\
    MMLU-de~\citep{lai-etal-2023-okapi}                         & de    & Knowledge                    & Accuracy & 1024 / 256 / 2048 & MIT \\
    INCLUDE-de~\citep{romanou2025include}                       & de    & Knowledge                    & Accuracy & 25 / 64 / 512     & Apache-2.0 \\
    MultiLoKo-de~\citep{hupkes2025multiloko}                    & de    & Knowledge                    & Accuracy & 16 / 0 / 234      & MIT \\
    \midrule
    WiC-ita~\citep{evalita2023}                                 & it    & Words in Context             & Macro F1 & 1024 / 256 / 1000 & --- \\
    MMLU-it~\citep{lai-etal-2023-okapi}                         & it    & Knowledge                    & Accuracy & 1024 / 256 / 2048 & MIT \\
    INCLUDE-it~\citep{romanou2025include}                       & it    & Knowledge                    & Accuracy & 25 / 64 / 512     & Apache-2.0 \\
    \midrule
    MLQA-es~\citep{lewis-etal-2020-mlqa}                        & es    & Knowledge                    & F1       & 1024 / 256 / 2048 & CC-BY-SA-3.0 \\
    INCLUDE-es~\citep{romanou2025include}                       & es    & Knowledge                    & Accuracy & 25 / 64 / 512     & Apache-2.0 \\
    MultiNRC-es~\citep{fabbri2025multinrc}                      & es    & Knowledge                    & Accuracy & 64 / 128 / 200    & --- \\
    MultiLoKo-es~\citep{hupkes2025multiloko}                    & es    & Knowledge                    & Accuracy & 16 / 0 / 234      & MIT \\
    \bottomrule
    \end{tabular}
    \caption{Evaluation datasets detailing their language, task category, measured metric, split sizes, and licensing. We do not make use of the train set.}
    \label{tab:evaluation_datasets}
\end{table*}

\section{Prompt Language}
\label{app:prompt_lang}
We ask whether the prompt language in the preference dataset matters, independent of the response language. We construct three variants of the multilingual DPO dataset: (\textbf{Chosen}) the prompt appears in the same language as the chosen response; (\textbf{Mixed}) prompts are assigned uniformly at random to one of the seven languages; (\textbf{Rejected}) the prompt appears in the same language as the rejected response, paired with a chosen response in a different language. Results for \eurollm{} are in~\cref{tab:dpo_prefix_ablations} (exact numbers are in~\cref{tab:dpo_prefix_ablations_per_dataset}).

The \textbf{Chosen} configuration is the strongest, it produces gains or ties in all languages except Italian, with the largest improvements on German ($+1.2$) and Danish ($+1.1$). The \textbf{Mixed} and \textbf{Rejected} variants degrade performance in five and six of seven languages respectively, with \textbf{Rejected} losing up to $4.7$ points on French. The Italian result, where \textbf{Mixed} ($+3.6$) and \textbf{Rejected} ($+7.3$) outperform \textbf{Chosen}, reflects the small number of Italian evaluation sets (4) and high variance within them. Overall, the prompt language should match the chosen response.

\section{Offline vs Online DPO}
\label{app:online_offline}

\begin{figure}[t]
    \centering
    \includegraphics[width=\linewidth]{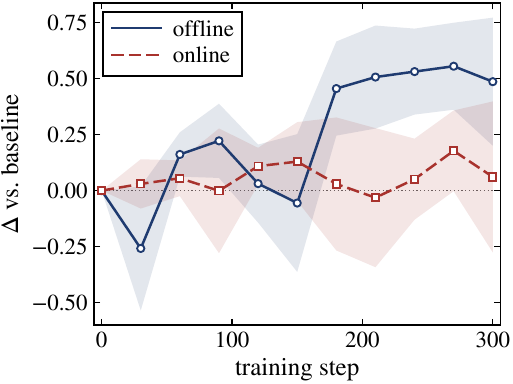}
    \caption{\textbf{Offline vs.\ online DPO on Danish evaluation tasks (\eurollm{}).} Average improvement over the baseline across 7 tasks; shaded regions denote standard deviation. Offline DPO converges to a higher plateau; online DPO is unstable and never exceeds $+0.2$ on average.}
    \label{fig:online_offline}
\end{figure}

We compare offline and online DPO directly. We adapt the work of \citet{guo2024directlanguagemodelalignment} to generate 16 responses (due to computational constraints) and score them with the same RM (Skywork). \cref{fig:online_offline} reports the average improvement over the baseline on the Danish tasks as a function of training step for \eurollm{}. Offline DPO reaches a peak of roughly $+0.6$ by step 200 and holds. Online DPO remains within $\pm 0.2$ of the baseline throughout training, with substantially higher variance.

Online DPO underperforms offline DPO when the RM is \emph{external} to the policy because online training creates a feedback loop: the policy optimizes against live RM scores on its own evolving output distribution, amplifying the biases the RM encodes and inducing exploitation of features the RM over-weights, rather than genuine preference learning. Offline DPO avoids this failure mode by treating the RM as a fixed labeler at dataset-construction time rather than a live optimization target, which decouples training-signal quality from the RM's reliability on the current policy's distribution. This matches \citet{pan2025what}, who show theoretically that online DPO reduces to SFT on the chosen responses.

\section{Additional m-ArenaHard 2.1 Results}\label{app:arenahard}

\begin{figure}[p]
    \centering
    \includegraphics[width=\linewidth]{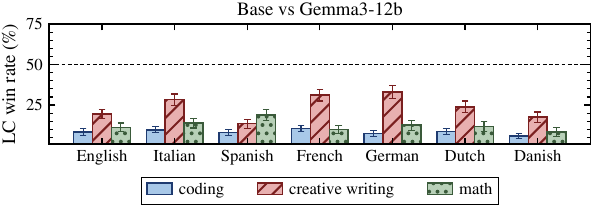}
    \caption{\textbf{m-ArenaHard 2.1 by subcategory: \eurollm{} base vs.\ Gemma3-12B-Instruct.} LC win rate broken down by prompt type. The \eurollm{} base loses across all categories and languages, with the largest deficits on math.}
    \label{fig:arenahard_subcat_base_vs_gemma_eurollm}
\end{figure}

\begin{figure}[p]
    \centering
    \includegraphics[width=\linewidth]{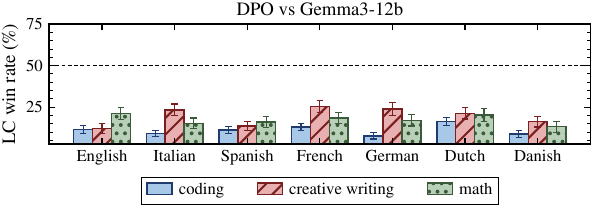}
    \caption{\textbf{m-ArenaHard 2.1 by subcategory: \eurollm{} {\tt Paired} DPO vs.\ Gemma3-12B-Instruct.} After DPO, win rates rise across most language-subcategory cells relative to~\cref{fig:arenahard_subcat_base_vs_gemma_eurollm}, with creative writing showing the most consistent improvement.}
    \label{fig:arenahard_subcat_dpo_vs_gemma_eurollm}
\end{figure}

\begin{figure}[p]
    \centering
    \includegraphics[width=\linewidth]{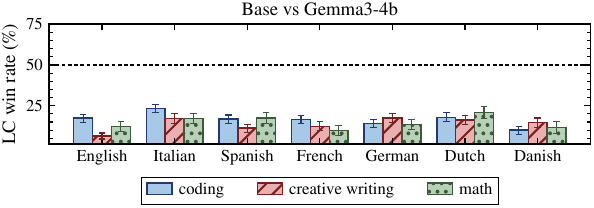}
    \caption{\textbf{m-ArenaHard 2.1 by subcategory: \aya{} base vs.\ Gemma3-4B-Instruct.} The \aya{} base loses across all subcategories.}
    \label{fig:arenahard_subcat_base_vs_gemma_aya}
\end{figure}

\begin{figure}[p]
    \centering
    \includegraphics[width=\linewidth]{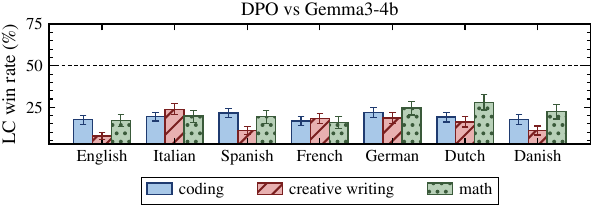}
    \caption{\textbf{m-ArenaHard 2.1 by subcategory: \aya{} {\tt Paired} DPO vs.\ Gemma3-4B-Instruct.} {\tt Paired} DPO improves all language-subcategory cells relative to~\cref{fig:arenahard_subcat_base_vs_gemma_aya}, with the largest gains on German and French.}
    \label{fig:arenahard_subcat_dpo_vs_gemma_aya}
\end{figure}

This appendix contains supplementary m-ArenaHard 2.1 figures referenced from~\cref{sec:arenahard}. \cref{fig:arenahard_subcat_base_vs_gemma_eurollm,fig:arenahard_subcat_dpo_vs_gemma_eurollm} show subcategory-level LC win rates against {\tt Gemma3-12B-Instruct} for \eurollm{}, before and after {\tt Paired} DPO. \cref{fig:arenahard_subcat_base_vs_gemma_aya,fig:arenahard_subcat_dpo_vs_gemma_aya} show the analogous comparison against {\tt Gemma3-4B-Instruct} for \aya{}.


\begin{table}[t]
    \centering
    \small
    \setlength{\tabcolsep}{3pt}
    \begin{tabular}{llcccc}
    \toprule
    & & \tt Baseline & \multicolumn{2}{c}{\tt SFT} & \multicolumn{1}{c}{\tt DPO} \\
    \cmidrule(lr){3-3} \cmidrule(lr){4-5} \cmidrule(lr){6-6}
    \multicolumn{2}{l}{Dataset} & {\tt EuroLLM} & {\tt All Lang} & {\tt Max-R} & {\tt Paired} \\
    \midrule
    \multirow{5}{*}{\rotatebox[origin=c]{90}{nor}} 
    & Idioms & 28.5 & \down{3.3} & \down{0.5} & \down{0.2} \\
    & MMLU & 49.4 & \down{6.9} & \down{0.4} & \up{0.1} \\
    & NorCS & 70.7 & \down{12.1} & 0.0 & \up{0.8} \\
    & NorQuAD & 74.2 & \down{1.3} & \up{0.6} & \up{1.0} \\
    & NRK Quiz & 48.6 & \down{4.7} & \up{0.9} & \up{0.6} \\
    \midrule
    \multirow{2}{*}{\rotatebox[origin=c]{90}{por}} 
    & MMLU & 53.4 & \down{10.8} & \down{0.2} & \down{0.2} \\
    & MultiLoKo & 41.3 & \down{5.5} & \up{0.3} & \up{1.0} \\
    \midrule
    \multirow{4}{*}{\rotatebox[origin=c]{90}{swe}} 
    & MMLU & 49.7 & \down{7.1} & \down{0.4} & \up{0.7} \\
    & MultiWikiQA & 77.6 & \down{1.4} & \up{0.6} & \up{1.9} \\
    & MultiLoKo & 36.8 & \down{8.8} & \down{2.7} & \down{1.7} \\
    & Skolprov & 47.2 & \down{3.5} & \up{1.2} & \down{0.1} \\
    \bottomrule
    \end{tabular}
    \caption{\textbf{Cross-lingual generalization to held-out languages in EuroEval (Norwegian, Portuguese, Swedish).} Values represent the absolute difference from the \eurollm{} baseline. {\tt Paired} DPO generalizes positively to 7/11 held-out datasets, while multilingual SFT catastrophically degrades performance on all of them.}
    \label{tab:eurollm_crosslingual_results}
\end{table}

\section{Per-Dataset Results}\label{app:per_dataset}

\cref{tab:per_dataset_aya,tab:per_dataset_eurollm} report per-dataset EuroEval scores for \aya{} and \eurollm{} respectively, across the same monolingual and multilingual configurations summarized in~\cref{tab:main_results}. \cref{tab:dpo_prefix_ablations_per_dataset} reports the per-dataset breakdown for the prompt-language ablation summarized in~\cref{tab:dpo_prefix_ablations}, \cref{tab:eurollm_en_vs_inlang_per_dataset} for the English-only vs.\ translated comparison summarized in~\cref{tab:eurollm_en_vs_inlang_diff}, and \cref{tab:eurollm_tinyaya_dpo_per_dataset} for the off-policy ablation summarized in~\cref{tab:eurollm_tinyaya_dpo_results}.

\begin{table*}[t]
    \centering
    \small
    \setlength{\tabcolsep}{4pt}
        \begin{tabular}{ll c cc c cc c}
    \toprule
    & & {\tt Baseline} & \multicolumn{2}{c}{\tt Monolingual SFT} & \multicolumn{1}{c}{\tt Mono.\ DPO} & \multicolumn{2}{c}{\tt Multilingual SFT} & \multicolumn{1}{c}{\tt Multi.\ DPO} \\
    \cmidrule(lr){3-3} \cmidrule(lr){4-5} \cmidrule(lr){6-6} \cmidrule(lr){7-8} \cmidrule(lr){9-9}
    \multicolumn{2}{l}{Dataset} & {\tt aya-3B} & {\tt In-lang} & {\tt Max-R} & {\tt Paired} & {\tt All Lang} & {\tt Max-R} & {\tt Paired} \\
    \midrule
    \multirow{7}{*}{\rotatebox[origin=c]{90}{dan}} & DaLA & 33.6 & 33.6 & 33.6 & 33.6 & 33.6 & 33.6 & 33.6 \\
     & Citizen & 62.5 & 61.6 & 57.0 & 62.5 & 58.0 & 55.8 & 62.5 \\
     & Entail. & 54.1 & 49.0 & 48.0 & 54.0 & 48.9 & 33.1 & 54.3 \\
     & LexInf. & 40.8 & 38.0 & 41.0 & 40.7 & 41.3 & 38.3 & 40.6 \\
     & Talemaader & 52.6 & 39.7 & 41.4 & 52.9 & 37.2 & 40.5 & 53.0 \\
     & DanWiC & 19.8 & 19.8 & 19.8 & 19.8 & 19.8 & 19.8 & 19.8 \\
     & MultiWikiQA & 75.1 & 68.5 & 68.3 & 75.7 & 67.7 & 65.6 & 75.5 \\
     & \textit{Avg.} & \textit{48.4} & \textit{44.3} & \textit{44.2} & \textit{48.5} & \textit{43.8} & \textit{41.0} & \textit{48.5} \\
    \midrule
    \multirow{4}{*}{\rotatebox[origin=c]{90}{deu}} & GermanQuAD & 58.3 & 54.5 & 50.8 & 58.2 & 52.7 & 48.3 & 58.4 \\
     & INCLUDE & 34.9 & 31.3 & 26.7 & 34.9 & 24.6 & 20.5 & 34.9 \\
     & MMLU & 49.5 & 38.9 & 41.3 & 49.5 & 36.9 & 41.3 & 49.5 \\
     & MultiLoKo & 27.0 & 28.3 & 28.0 & 26.5 & 28.0 & 29.0 & 26.5 \\
     & \textit{Avg.} & \textit{42.4} & \textit{38.2} & \textit{36.7} & \textit{42.3} & \textit{35.5} & \textit{34.8} & \textit{42.3} \\
    \midrule
    \multirow{5}{*}{\rotatebox[origin=c]{90}{eng}} & LifeInUK & 71.2 & 69.4 & 66.0 & 71.6 & 64.1 & 65.0 & 71.5 \\
     & MMLU-Pro & 27.9 & 23.3 & 22.4 & 28.0 & 21.3 & 22.6 & 28.4 \\
     & MultiLoKo & 33.3 & 30.2 & 31.6 & 33.6 & 34.3 & 32.9 & 34.2 \\
     & SQuAD & 82.4 & 80.0 & 75.2 & 82.5 & 77.7 & 74.2 & 82.3 \\
     & WiC & 55.9 & 53.7 & 43.9 & 56.0 & 33.8 & 41.2 & 56.4 \\
     & \textit{Avg.} & \textit{54.1} & \textit{51.3} & \textit{47.8} & \textit{54.3} & \textit{46.2} & \textit{47.2} & \textit{54.6} \\
    \midrule
    \multirow{4}{*}{\rotatebox[origin=c]{90}{spa}} & INCLUDE & 51.9 & 44.4 & 44.7 & 52.9 & 41.1 & 44.9 & 52.2 \\
     & MLQA & 65.4 & 58.9 & 57.9 & 65.2 & 58.3 & 56.9 & 65.2 \\
     & MultiLoKo & 27.4 & 32.0 & 30.1 & 26.5 & 28.5 & 28.1 & 26.4 \\
     & MultiNRC & 26.4 & 30.1 & 26.6 & 26.2 & 27.8 & 23.1 & 25.4 \\
     & \textit{Avg.} & \textit{42.8} & \textit{41.4} & \textit{39.8} & \textit{42.7} & \textit{38.9} & \textit{38.2} & \textit{42.3} \\
    \midrule
    \multirow{5}{*}{\rotatebox[origin=c]{90}{fra}} & FQuAD & 70.5 & 64.5 & 63.0 & 70.5 & 62.5 & 59.4 & 70.6 \\
     & INCLUDE & 49.6 & 44.9 & 43.2 & 49.6 & 37.2 & 43.4 & 49.5 \\
     & MMLU & 48.8 & 41.2 & 41.8 & 48.8 & 39.1 & 40.8 & 48.9 \\
     & MultiLoKo & 26.6 & 23.6 & 20.7 & 26.2 & 24.6 & 19.7 & 25.4 \\
     & MultiNRC & 34.0 & 29.2 & 27.1 & 34.3 & 30.3 & 30.3 & 33.6 \\
     & \textit{Avg.} & \textit{45.9} & \textit{40.7} & \textit{39.2} & \textit{45.9} & \textit{38.7} & \textit{38.7} & \textit{45.6} \\
    \midrule
    \multirow{3}{*}{\rotatebox[origin=c]{90}{ita}} & INCLUDE & 55.5 & 47.3 & 46.0 & 56.2 & 44.3 & 44.8 & 56.5 \\
     & MMLU & 48.0 & 40.3 & 40.8 & 48.1 & 38.9 & 41.3 & 47.7 \\
     & WiC & 51.5 & 35.3 & 36.4 & 51.8 & 38.0 & 37.4 & 52.7 \\
     & \textit{Avg.} & \textit{51.7} & \textit{41.0} & \textit{41.1} & \textit{52.0} & \textit{40.4} & \textit{41.2} & \textit{52.3} \\
    \midrule
    \multirow{4}{*}{\rotatebox[origin=c]{90}{nld}} & COPA & 82.4 & 76.3 & 79.4 & 82.7 & 58.9 & 74.1 & 83.1 \\
     & INCLUDE & 52.2 & 42.5 & 45.6 & 51.6 & 44.1 & 46.5 & 51.7 \\
     & MultiLoKo & 25.3 & 24.8 & 20.2 & 25.3 & 24.3 & 21.9 & 25.9 \\
     & SQuAD & 74.0 & 66.3 & 64.3 & 74.0 & 63.6 & 62.1 & 74.2 \\
     & \textit{Avg.} & \textit{58.5} & \textit{52.5} & \textit{52.4} & \textit{58.4} & \textit{47.7} & \textit{51.1} & \textit{58.7} \\
    \bottomrule
    \end{tabular}

    \caption{\textbf{Per-dataset EuroEval scores for \texttt{aya-3B}.} All values are absolute scores (averaged over three seeds). Columns under {\tt Monolingual} use post-training data only in the row-language; columns under {\tt Multilingual} pool data across all seven languages. {\tt Max-R} denotes filtering by the maximum-reward response from a pool of candidates; {\tt Paired} denotes DPO using chosen/rejected pairs.}
    \label{tab:per_dataset_aya}
\end{table*}

\begin{table*}[t]
    \centering
    \small
    \setlength{\tabcolsep}{4pt}
        \begin{tabular}{ll c cc c cc c}
    \toprule
    & & {\tt Baseline} & \multicolumn{2}{c}{\tt Monolingual SFT} & \multicolumn{1}{c}{\tt Mono.\ DPO} & \multicolumn{2}{c}{\tt Multilingual SFT} & \multicolumn{1}{c}{\tt Multi.\ DPO} \\
    \cmidrule(lr){3-3} \cmidrule(lr){4-5} \cmidrule(lr){6-6} \cmidrule(lr){7-8} \cmidrule(lr){9-9}
    \multicolumn{2}{l}{Dataset} & {\tt EuroLLM-9B} & {\tt In-lang} & {\tt Max-R} & {\tt Paired} & {\tt All Lang} & {\tt Max-R} & {\tt Paired} \\
    \midrule
    \multirow{7}{*}{\rotatebox[origin=c]{90}{dan}} & DaLA & 48.2 & 37.6 & 43.0 & 49.8 & 41.2 & 50.2 & 51.4 \\
     & Citizen & 81.4 & 81.1 & 88.2 & 82.2 & 77.1 & 83.4 & 81.6 \\
     & Entail. & 67.5 & 59.6 & 71.0 & 68.2 & 56.8 & 68.6 & 67.3 \\
     & LexInf. & 76.2 & 73.2 & 68.5 & 77.3 & 61.9 & 76.4 & 76.5 \\
     & Talemaader & 57.7 & 52.4 & 56.8 & 57.9 & 47.4 & 58.5 & 57.6 \\
     & DanWiC & 45.0 & 34.7 & 50.0 & 45.6 & 34.0 & 47.1 & 47.2 \\
     & MultiWikiQA & 79.4 & 80.0 & 78.4 & 81.0 & 77.4 & 80.2 & 81.3 \\
     & \textit{Avg.} & \textit{65.1} & \textit{59.8} & \textit{65.1} & \textit{66.0} & \textit{56.5} & \textit{66.3} & \textit{66.1} \\
    \midrule
    \multirow{4}{*}{\rotatebox[origin=c]{90}{deu}} & GermanQuAD & 57.4 & 63.9 & 60.7 & 59.8 & 59.4 & 60.3 & 60.2 \\
     & INCLUDE & 36.9 & 44.1 & 34.9 & 38.5 & 38.0 & 41.7 & 39.5 \\
     & MMLU & 53.2 & 49.8 & 46.9 & 53.7 & 45.9 & 45.7 & 53.2 \\
     & MultiLoKo & 41.9 & 33.3 & 39.1 & 41.6 & 30.9 & 38.3 & 41.2 \\
     & \textit{Avg.} & \textit{47.4} & \textit{47.8} & \textit{45.4} & \textit{48.4} & \textit{43.5} & \textit{46.5} & \textit{48.5} \\
    \midrule
    \multirow{5}{*}{\rotatebox[origin=c]{90}{eng}} & LifeInUK & 75.0 & 77.5 & 76.6 & 75.1 & 73.1 & 77.2 & 75.8 \\
     & MMLU-Pro & 28.4 & 24.4 & 25.9 & 27.9 & 21.8 & 27.4 & 28.4 \\
     & MultiLoKo & 33.3 & 34.5 & 34.2 & 30.8 & 31.6 & 34.3 & 32.5 \\
     & SQuAD & 82.8 & 86.0 & 82.6 & 83.7 & 82.5 & 83.7 & 83.9 \\
     & WiC & 64.5 & 61.0 & 62.1 & 64.8 & 55.5 & 62.6 & 65.2 \\
     & \textit{Avg.} & \textit{56.8} & \textit{56.7} & \textit{56.3} & \textit{56.5} & \textit{52.9} & \textit{57.0} & \textit{57.2} \\
    \midrule
    \multirow{4}{*}{\rotatebox[origin=c]{90}{spa}} & INCLUDE & 63.4 & 59.4 & 59.9 & 53.1 & 54.6 & 55.2 & 64.3 \\
     & MLQA & 66.0 & 69.6 & 69.6 & 67.9 & 65.6 & 67.5 & 66.3 \\
     & MultiLoKo & 42.9 & 37.2 & 40.0 & 43.2 & 34.8 & 41.2 & 43.7 \\
     & MultiNRC & 34.9 & 29.2 & 30.7 & 37.8 & 33.0 & 34.6 & 36.2 \\
     & \textit{Avg.} & \textit{51.8} & \textit{48.9} & \textit{50.0} & \textit{50.5} & \textit{47.0} & \textit{49.6} & \textit{52.6} \\
    \midrule
    \multirow{5}{*}{\rotatebox[origin=c]{90}{fra}} & FQuAD & 68.4 & 72.1 & 71.9 & 70.3 & 68.4 & 71.3 & 70.4 \\
     & INCLUDE & 64.9 & 60.0 & 55.2 & 64.6 & 55.1 & 54.7 & 64.9 \\
     & MMLU & 53.8 & 52.1 & 50.5 & 53.7 & 46.2 & 48.4 & 54.1 \\
     & MultiLoKo & 33.5 & 30.1 & 34.8 & 36.0 & 29.2 & 32.6 & 35.8 \\
     & MultiNRC & 40.3 & 36.3 & 41.9 & 39.1 & 28.9 & 40.1 & 39.8 \\
     & \textit{Avg.} & \textit{52.2} & \textit{50.1} & \textit{50.9} & \textit{52.7} & \textit{45.6} & \textit{49.4} & \textit{53.0} \\
    \midrule
    \multirow{3}{*}{\rotatebox[origin=c]{90}{ita}} & INCLUDE & 58.7 & 53.4 & 59.9 & 59.4 & 50.1 & 58.9 & 59.1 \\
     & MMLU & 52.6 & 48.5 & 58.5 & 62.0 & 42.6 & 61.3 & 52.4 \\
     & WiC & 51.7 & 38.2 & 48.4 & 52.3 & 40.5 & 57.7 & 47.6 \\
     & \textit{Avg.} & \textit{54.3} & \textit{46.7} & \textit{55.6} & \textit{57.9} & \textit{44.4} & \textit{59.3} & \textit{53.0} \\
    \midrule
    \multirow{4}{*}{\rotatebox[origin=c]{90}{nld}} & COPA & 93.5 & 92.0 & 93.7 & 93.3 & 88.1 & 93.5 & 93.0 \\
     & INCLUDE & 62.5 & 59.5 & 51.6 & 62.2 & 57.3 & 61.7 & 61.0 \\
     & MultiLoKo & 41.9 & 29.8 & 39.5 & 41.3 & 26.9 & 36.9 & 41.3 \\
     & SQuAD & 74.1 & 75.2 & 80.5 & 76.2 & 73.6 & 76.1 & 76.8 \\
     & \textit{Avg.} & \textit{68.0} & \textit{64.1} & \textit{66.3} & \textit{68.2} & \textit{61.5} & \textit{67.0} & \textit{68.0} \\
    \bottomrule
    \end{tabular}

    \caption{\textbf{Per-dataset EuroEval scores for \texttt{EuroLLM-9B-Instruct}.} All values are absolute scores (averaged over three seeds). Columns under {\tt Monolingual} use post-training data only in the row-language; columns under {\tt Multilingual} pool data across all seven languages. {\tt Max-R} denotes filtering by the maximum-reward response from a pool of candidates; {\tt Paired} denotes DPO using chosen/rejected pairs.}
    \label{tab:per_dataset_eurollm}
\end{table*}

\section{Per Dataset Numbers for Held-out Set}
\label{app:exact_nums}
In~\cref{tab:per_dataset_eurollm}, \cref{tab:per_dataset_aya}, and \cref{tab:eurollm_crosslingual_results}, we show the exact per-dataset evaluation results for \aya{}, \eurollm{}, and cross-lingually to Norwegian, Portuguese, and Swedish.

\section{Large Language Model Use}
We made use of LLM harnesses agents (namely Claude) to polish our writing, coding to an extent, and plotting our figures.

\begin{table*}[t]
    \centering
    \small
    \setlength{\tabcolsep}{4pt}
    \begin{tabular}{ll cccc}
    \toprule
    & & {\tt Baseline} & \multicolumn{3}{c}{\tt Language of the Prompt} \\
    \cmidrule(lr){3-3} \cmidrule(lr){4-6}
    \multicolumn{2}{l}{Dataset} & {\tt 9B-Inst} & {\tt Chosen} & {\tt Mixed} & {\tt Rejected} \\
    \midrule
    \multirow{8}{*}{\rotatebox[origin=c]{90}{dan}} & DaLA & 48.2 & 51.4 & 45.4 & 39.7 \\
     & Citizen & 81.4 & 81.6 & 87.0 & 87.4 \\
     & Entail. & 67.5 & 67.3 & 66.5 & 66.0 \\
     & LexInf. & 76.2 & 76.5 & 68.3 & 70.8 \\
     & Talemaader & 57.7 & 57.6 & 55.7 & 53.1 \\
     & DanWiC & 45.0 & 47.2 & 49.5 & 49.6 \\
     & MultiWikiQA & 79.4 & 81.3 & 81.3 & 78.8 \\
     & \textit{Avg.} & \textit{65.1} & \textit{66.1} & \textit{64.8} & \textit{63.6} \\
    \midrule
    \multirow{5}{*}{\rotatebox[origin=c]{90}{deu}} & GermanQuAD & 57.4 & 60.2 & 59.1 & 56.5 \\
     & INCLUDE & 36.9 & 39.5 & 35.9 & 38.0 \\
     & MMLU & 53.2 & 53.2 & 48.7 & 48.4 \\
     & MultiLoKo & 41.9 & 41.2 & 43.5 & 42.1 \\
     & \textit{Avg.} & \textit{47.4} & \textit{48.5} & \textit{46.8} & \textit{46.2} \\
    \midrule
    \multirow{6}{*}{\rotatebox[origin=c]{90}{eng}} & LifeInUK & 75.0 & 75.8 & 76.7 & 76.4 \\
     & MMLU-Pro & 28.4 & 28.4 & 26.4 & 27.3 \\
     & MultiLoKo & 33.3 & 32.5 & 31.6 & 30.6 \\
     & SQuAD & 82.8 & 83.9 & 83.2 & 81.7 \\
     & WiC & 64.5 & 65.2 & 58.7 & 61.3 \\
     & \textit{Avg.} & \textit{56.8} & \textit{57.2} & \textit{55.3} & \textit{55.5} \\
    \midrule
    \multirow{5}{*}{\rotatebox[origin=c]{90}{spa}} & INCLUDE & 63.4 & 64.3 & 53.1 & 52.6 \\
     & MLQA & 66.0 & 66.3 & 68.2 & 66.6 \\
     & MultiLoKo & 42.9 & 43.7 & 44.3 & 41.9 \\
     & MultiNRC & 34.9 & 36.2 & 39.3 & 38.0 \\
     & \textit{Avg.} & \textit{51.8} & \textit{52.6} & \textit{51.2} & \textit{49.8} \\
    \midrule
    \multirow{6}{*}{\rotatebox[origin=c]{90}{fra}} & FQuAD & 68.4 & 70.4 & 69.6 & 67.0 \\
     & INCLUDE & 64.9 & 64.9 & 53.1 & 51.6 \\
     & MMLU & 53.8 & 54.1 & 48.6 & 47.9 \\
     & MultiLoKo & 33.5 & 35.8 & 35.2 & 32.0 \\
     & MultiNRC & 40.3 & 39.8 & 41.9 & 38.8 \\
     & \textit{Avg.} & \textit{52.2} & \textit{53.0} & \textit{49.7} & \textit{47.5} \\
    \midrule
    \multirow{4}{*}{\rotatebox[origin=c]{90}{ita}} & INCLUDE & 58.7 & 59.1 & 60.4 & 57.8 \\
     & MMLU & 52.6 & 52.4 & 60.3 & 61.3 \\
     & WiC & 51.7 & 47.6 & 53.0 & 65.7 \\
     & \textit{Avg.} & \textit{54.3} & \textit{53.0} & \textit{57.9} & \textit{61.6} \\
    \midrule
    \multirow{5}{*}{\rotatebox[origin=c]{90}{nld}} & COPA & 93.5 & 93.0 & 92.0 & 90.7 \\
     & INCLUDE & 62.5 & 61.0 & 48.4 & 54.2 \\
     & MultiLoKo & 41.9 & 41.3 & 41.6 & 40.5 \\
     & SQuAD & 74.1 & 76.8 & 79.5 & 75.3 \\
     & \textit{Avg.} & \textit{68.0} & \textit{68.0} & \textit{65.4} & \textit{65.2} \\
    \bottomrule
    \end{tabular}
    \caption{\textbf{Per-dataset DPO prompt-language ablation (\eurollm{}).} All values are absolute scores (averaged over three seeds). The italicized {\it Avg.\ } row at the end of each language block reports the language-level mean. {\tt Chosen} pairs the prompt with the same-language chosen response; {\tt Mixed} samples prompt language uniformly; {\tt Rejected} pairs the prompt with the same-language rejected response.}
    \label{tab:dpo_prefix_ablations_per_dataset}
\end{table*}

\begin{table*}[t]
    \centering
    \small
    \setlength{\tabcolsep}{4pt}
    \begin{tabular}{ll c cc cc cc}
    \toprule
    & & {\tt Baseline} & \multicolumn{2}{c}{\tt SFT} & \multicolumn{2}{c}{\tt Max-R} & \multicolumn{2}{c}{\tt Paired} \\
    \cmidrule(lr){3-3} \cmidrule(lr){4-5} \cmidrule(lr){6-7} \cmidrule(lr){8-9}
    \multicolumn{2}{l}{Dataset} & {\tt 9B-Inst} & {\tt eng} & {\tt tgt} & {\tt eng} & {\tt tgt} & {\tt eng} & {\tt tgt} \\
    \midrule
    \multirow{8}{*}{\rotatebox[origin=c]{90}{dan}} & DaLA & 48.2 & 40.0 & 37.6 & 47.5 & 43.0 & 48.5 & 49.8 \\
     & Citizen & 81.4 & 82.6 & 81.1 & 81.8 & 88.2 & 81.4 & 82.2 \\
     & Entail. & 67.5 & 63.1 & 59.6 & 68.5 & 71.0 & 67.8 & 68.2 \\
     & LexInf. & 76.2 & 75.9 & 73.2 & 76.7 & 68.5 & 76.2 & 77.3 \\
     & Talemaader & 57.7 & 57.7 & 52.4 & 58.8 & 56.8 & 57.0 & 57.9 \\
     & DanWiC & 45.0 & 46.2 & 34.7 & 54.0 & 50.0 & 56.3 & 45.6 \\
     & MultiWikiQA & 79.4 & 80.9 & 80.0 & 80.4 & 78.4 & 80.5 & 81.0 \\
     & \textit{Avg.} & \textit{65.0} & \textit{63.8} & \textit{59.8} & \textit{66.8} & \textit{65.1} & \textit{66.8} & \textit{66.0} \\
    \midrule
    \multirow{5}{*}{\rotatebox[origin=c]{90}{deu}} & GermanQuAD & 57.4 & 63.7 & 63.9 & 61.6 & 60.7 & 58.9 & 59.8 \\
     & INCLUDE & 36.9 & 43.1 & 44.1 & 39.5 & 34.9 & 38.5 & 38.5 \\
     & MMLU & 53.2 & 50.8 & 49.8 & 52.6 & 46.9 & 53.6 & 53.7 \\
     & MultiLoKo & 41.9 & 34.8 & 33.3 & 37.3 & 39.1 & 42.2 & 41.6 \\
     & \textit{Avg.} & \textit{47.4} & \textit{48.1} & \textit{47.8} & \textit{47.8} & \textit{45.4} & \textit{48.3} & \textit{48.4} \\
    \midrule
    \multirow{5}{*}{\rotatebox[origin=c]{90}{spa}} & INCLUDE & 63.4 & 58.9 & 59.4 & 62.9 & 59.9 & 64.1 & 53.1 \\
     & MLQA & 66.0 & 68.7 & 69.6 & 67.1 & 69.6 & 66.4 & 67.9 \\
     & MultiLoKo & 42.9 & 38.9 & 37.2 & 39.9 & 40.0 & 42.9 & 43.2 \\
     & MultiNRC & 34.9 & 36.8 & 29.2 & 38.5 & 30.7 & 34.7 & 37.8 \\
     & \textit{Avg.} & \textit{51.8} & \textit{50.8} & \textit{48.8} & \textit{52.1} & \textit{50.1} & \textit{52.0} & \textit{50.5} \\
    \midrule
    \multirow{6}{*}{\rotatebox[origin=c]{90}{fra}} & FQuAD & 68.4 & 72.0 & 72.1 & 72.5 & 71.9 & 69.6 & 70.3 \\
     & INCLUDE & 64.9 & 59.8 & 60.0 & 65.3 & 55.2 & 65.0 & 64.6 \\
     & MMLU & 53.8 & 52.2 & 52.1 & 53.6 & 50.5 & 54.2 & 53.7 \\
     & MultiLoKo & 33.5 & 30.3 & 30.1 & 34.0 & 34.8 & 33.0 & 36.0 \\
     & MultiNRC & 40.3 & 33.1 & 36.3 & 36.6 & 41.9 & 40.0 & 39.1 \\
     & \textit{Avg.} & \textit{52.2} & \textit{49.5} & \textit{50.1} & \textit{52.4} & \textit{50.9} & \textit{52.4} & \textit{52.8} \\
    \midrule
    \multirow{4}{*}{\rotatebox[origin=c]{90}{ita}} & INCLUDE & 58.7 & 55.0 & 53.4 & 57.5 & 59.9 & 58.4 & 59.4 \\
     & MMLU & 52.6 & 50.7 & 48.5 & 51.9 & 58.5 & 52.4 & 62.0 \\
     & WiC & 51.7 & 43.2 & 38.2 & 44.9 & 48.4 & 50.2 & 52.3 \\
     & \textit{Avg.} & \textit{54.3} & \textit{49.7} & \textit{46.7} & \textit{51.4} & \textit{55.6} & \textit{53.7} & \textit{57.9} \\
    \midrule
    \multirow{5}{*}{\rotatebox[origin=c]{90}{nld}} & COPA & 93.5 & 92.9 & 92.0 & 93.1 & 93.7 & 93.1 & 93.3 \\
     & INCLUDE & 62.5 & 61.2 & 59.5 & 62.0 & 51.6 & 61.8 & 62.2 \\
     & MultiLoKo & 41.9 & 33.9 & 29.8 & 37.9 & 39.5 & 41.3 & 41.3 \\
     & SQuAD & 74.1 & 77.6 & 75.2 & 76.3 & 80.5 & 75.5 & 76.2 \\
     & \textit{Avg.} & \textit{68.0} & \textit{66.4} & \textit{64.1} & \textit{67.3} & \textit{66.3} & \textit{67.9} & \textit{68.2} \\
    \bottomrule
    \end{tabular}
    \caption{\textbf{Per-dataset English-only vs.\ translated in-language post-training (\eurollm{}).} All values are absolute scores (averaged over three seeds). The italicized {\it Avg.\ } row at the end of each language block reports the language-level mean. {\tt eng} columns are post-training using English-only data; {\tt tgt} columns are post-training using the row-language data. {\tt SFT}, {\tt Max-R}, and {\tt Paired} correspond to the same monolingual interventions defined in the main results.}
    \label{tab:eurollm_en_vs_inlang_per_dataset}
\end{table*}

\begin{table*}[t]
    \centering
    \small
    \setlength{\tabcolsep}{4pt}
    \begin{tabular}{ll c ccc ccc}
    \toprule
    & & {\tt Base} & \multicolumn{3}{c}{\tt Mono. PT} & \multicolumn{3}{c}{\tt Multi. PT} \\
    \cmidrule(lr){3-3} \cmidrule(lr){4-6} \cmidrule(lr){7-9}
    \multicolumn{2}{l}{Dataset} & {\tt 9B-Inst} & {\tt In SFT} & {\tt Max SFT} & {\tt Pair DPO} & {\tt All SFT} & {\tt Max SFT} & {\tt Pair DPO} \\
    \midrule
    \multirow{8}{*}{\rotatebox[origin=c]{90}{dan}} & DaLA & 48.2 & 37.6 & 38.2 & 48.8 & 41.2 & 46.6 & 48.3 \\
     & Citizen & 81.4 & 81.1 & 73.2 & 81.4 & 77.1 & 73.9 & 81.5 \\
     & Entail. & 67.5 & 59.6 & 58.4 & 67.0 & 56.8 & 57.4 & 67.1 \\
     & LexInf. & 76.2 & 73.2 & 61.5 & 77.1 & 61.9 & 63.1 & 76.8 \\
     & Talemaader & 57.7 & 52.4 & 44.2 & 57.7 & 47.4 & 39.5 & 58.2 \\
     & DanWiC & 45.0 & 34.7 & 49.9 & 56.0 & 34.0 & 52.9 & 56.0 \\
     & MultiWikiQA & 79.4 & 80.0 & 76.7 & 79.1 & 77.4 & 77.7 & 79.1 \\
     & \textit{Avg.} & \textit{65.0} & \textit{59.8} & \textit{57.5} & \textit{66.7} & \textit{56.5} & \textit{58.7} & \textit{66.7} \\
    \midrule
    \multirow{5}{*}{\rotatebox[origin=c]{90}{deu}} & GermanQuAD & 57.4 & 63.9 & 62.4 & 57.0 & 59.4 & 60.0 & 57.2 \\
     & INCLUDE & 36.9 & 44.1 & 36.9 & 38.0 & 38.0 & 42.0 & 38.0 \\
     & MMLU & 53.2 & 49.8 & 45.1 & 53.6 & 45.9 & 42.4 & 53.4 \\
     & MultiLoKo & 41.9 & 33.3 & 32.3 & 42.2 & 30.9 & 31.6 & 43.1 \\
     & \textit{Avg.} & \textit{47.4} & \textit{47.8} & \textit{44.2} & \textit{47.7} & \textit{43.5} & \textit{44.0} & \textit{47.9} \\
    \midrule
    \multirow{6}{*}{\rotatebox[origin=c]{90}{eng}} & LifeInUK & 75.0 & 77.5 & 72.7 & 75.7 & 73.1 & 68.0 & 75.5 \\
     & MMLU-Pro & 28.4 & 24.4 & 23.2 & 28.1 & 21.8 & 20.3 & 28.1 \\
     & MultiLoKo & 33.3 & 34.5 & 31.6 & 31.3 & 31.6 & 25.7 & 31.5 \\
     & SQuAD & 82.8 & 86.0 & 83.1 & 82.2 & 82.5 & 82.7 & 82.8 \\
     & WiC & 64.5 & 61.0 & 61.2 & 63.9 & 55.5 & 57.4 & 64.2 \\
     & \textit{Avg.} & \textit{56.8} & \textit{56.7} & \textit{54.4} & \textit{56.2} & \textit{52.9} & \textit{50.8} & \textit{56.4} \\
    \midrule
    \multirow{5}{*}{\rotatebox[origin=c]{90}{spa}} & INCLUDE & 63.4 & 59.4 & 54.6 & 63.7 & 54.6 & 49.7 & 64.2 \\
     & MLQA & 66.0 & 69.6 & 66.9 & 65.8 & 65.6 & 64.5 & 65.7 \\
     & MultiLoKo & 42.9 & 37.2 & 31.9 & 43.9 & 34.8 & 32.9 & 43.2 \\
     & MultiNRC & 34.9 & 29.2 & 38.1 & 34.5 & 33.0 & 34.3 & 35.3 \\
     & \textit{Avg.} & \textit{51.8} & \textit{48.8} & \textit{47.9} & \textit{52.0} & \textit{47.0} & \textit{45.4} & \textit{52.1} \\
    \midrule
    \multirow{6}{*}{\rotatebox[origin=c]{90}{fra}} & FQuAD & 68.4 & 72.1 & 71.3 & 68.7 & 68.4 & 68.4 & 68.5 \\
     & INCLUDE & 64.9 & 60.0 & 52.7 & 64.3 & 55.1 & 51.6 & 64.6 \\
     & MMLU & 53.8 & 52.1 & 48.1 & 54.0 & 46.2 & 47.6 & 54.0 \\
     & MultiLoKo & 33.5 & 30.1 & 25.9 & 34.5 & 29.2 & 23.5 & 34.0 \\
     & MultiNRC & 40.3 & 36.3 & 35.0 & 40.0 & 28.9 & 32.4 & 40.0 \\
     & \textit{Avg.} & \textit{52.2} & \textit{50.1} & \textit{46.6} & \textit{52.3} & \textit{45.6} & \textit{44.7} & \textit{52.2} \\
    \midrule
    \multirow{4}{*}{\rotatebox[origin=c]{90}{ita}} & INCLUDE & 58.7 & 53.4 & 52.7 & 58.1 & 50.1 & 47.7 & 58.6 \\
     & MMLU & 52.6 & 48.5 & 45.0 & 52.4 & 42.6 & 44.0 & 52.3 \\
     & WiC & 51.7 & 38.2 & 51.5 & 51.1 & 40.5 & 49.2 & 51.7 \\
     & \textit{Avg.} & \textit{54.3} & \textit{46.7} & \textit{49.8} & \textit{53.9} & \textit{44.4} & \textit{47.0} & \textit{54.2} \\
    \midrule
    \multirow{5}{*}{\rotatebox[origin=c]{90}{nld}} & COPA & 93.5 & 92.0 & 88.3 & 93.4 & 88.1 & 76.7 & 93.0 \\
     & INCLUDE & 62.5 & 59.5 & 56.0 & 61.7 & 57.3 & 50.7 & 61.8 \\
     & MultiLoKo & 41.9 & 29.8 & 32.0 & 40.8 & 26.9 & 30.2 & 40.5 \\
     & SQuAD & 74.1 & 75.2 & 75.0 & 73.6 & 73.6 & 73.3 & 73.9 \\
     & \textit{Avg.} & \textit{68.0} & \textit{64.1} & \textit{62.8} & \textit{67.4} & \textit{61.5} & \textit{57.7} & \textit{67.3} \\
    \bottomrule
    \end{tabular}
    \caption{\textbf{Per-dataset off-policy data ablation: tuning \eurollm{} on \aya{} generations.} All values are absolute scores (averaged over three seeds). The italicized {\it Avg.\ } row at the end of each language block reports the language-level mean. The {\tt Max SFT} and {\tt Pair DPO} columns under both {\tt Mono. PT} and {\tt Multi. PT} use post-training data generated by {\tt aya-3B}; for reference, the {\tt In SFT} and {\tt All SFT} columns reuse the on-policy translated-data SFT runs from \cref{tab:main_results}.}
    \label{tab:eurollm_tinyaya_dpo_per_dataset}
\end{table*}

\end{document}